\documentclass[pdflatex, sn-nature, oneside]{sn-jnl}

\makeatletter
\def\headerps@out{}
\makeatother

\usepackage{graphicx}%
\usepackage{multirow}%
\usepackage{amsmath,amssymb,amsfonts}%
\usepackage{amsthm}%
\usepackage{mathrsfs}%
\usepackage[title]{appendix}%
\usepackage[table]{xcolor}%
\usepackage{textcomp}%
\usepackage{manyfoot}%
\usepackage{tabularx}
\usepackage{caption}
\usepackage{enumitem}
\usepackage{booktabs}%
\usepackage{algorithm}%
\usepackage{algorithmicx}%
\usepackage{algpseudocode}%
\usepackage{listings}%
\usepackage[most]{tcolorbox}
\usepackage[skip=0.5\baselineskip]{parskip}
\usepackage{makecell}
\usepackage{newunicodechar}
\usepackage{wasysym}
\usepackage[normalem]{ulem}
\usepackage{relsize}
\usepackage{lineno}
\usepackage{xltabular}
\usepackage{array}

\theoremstyle{thmstyleone}%
\theoremstyle{thmstyletwo}%

\theoremstyle{thmstylethree}%

\lstdefinestyle{promptcontent}{
    basicstyle=\ttfamily\small,
    breaklines=true,
    breakindent=0pt,        
    breakatwhitespace=false, 
    columns=fullflexible,
    keepspaces=true,
}

\DeclareCaptionLabelSeparator{vline}{ $|$ } 

\newcommand{\IC}{\operatorname{IC}}

\newcommand{\Sim}{\operatorname{Sim}}

\newcommand{\Anc}{\operatorname{Anc}}

\newunicodechar{●}{\ensuremath{\CIRCLE}}
\newunicodechar{○}{\ensuremath{\Circle}}
\newunicodechar{◐}{\ensuremath{\LEFTcircle}}

\definecolor{tablegray}{gray}{0.92}
\definecolor{yellowheader}{HTML}{FFE699} 
\definecolor{yellowrow}{HTML}{FFF2CC}    
\definecolor{blueheader}{HTML}{BDD7EE}   
\definecolor{bluerow}{HTML}{DDEBF7}      

\renewcommand{\abstractname}{}

\tcbuselibrary{skins, breakable} 

\newtcolorbox{promptbox}[1]{%
    colback=gray!5,             
    colframe=gray!60,           
    coltitle=white,             
    colbacktitle=gray!60,       
    fonttitle=\bfseries\small,  
    fontupper=\sffamily\small,  
    arc=0mm,                    
    boxrule=0.8pt,              
    title=#1,                   
    breakable,                  
    enhanced,                   
    before=\vspace{1em},        
    after=\vspace{1em}          
}

\setcitestyle{super,open={},close={}}

\begin{document}
\title[]{\vspace{-0em}\textbf{Auditing evaluation artifacts and data leakage in large language model systems for rare disease diagnosis}\vspace{-1em}}

\author{
    \relsize{-0.5}
    \fnm{Jian} \sur{Yang}\textsuperscript{1,*}, 
    \fnm{Peilin} \sur{Liu}\textsuperscript{2}, 
    \fnm{Xiaohan} \sur{Li}\textsuperscript{2},
    \fnm{Songda} \sur{Li}\textsuperscript{2}, 
    \fnm{Yu} \sur{Cao}\textsuperscript{1},
    \fnm{Yuqing} \sur{Feng}\textsuperscript{1},
    \fnm{and Haomin} \sur{Li}\textsuperscript{1,2,*}
    \vspace{-0.5em}
}

\affil[1]{\small \orgdiv{The Children's Hospital}, \orgname{Zhejiang University School of Medicine}, \orgaddress{\city{Hangzhou}, \postcode{310052}, \country{China}}\vspace{-0.3em}}

\affil[2]{\small \orgdiv{The College of Biomedical Engineering and Instrument Science}, \orgname{Zhejiang University}, \orgaddress{\city{Hangzhou}, \postcode{310027}, \country{China}}\vspace{-0.3em}}

\affil[]{\small \textsuperscript{*}Corresponding Author: hmli@zju.edu.cn (H. L.); yangjian\_bme@zju.edu.cn (J. Y.)\vspace{-4em}}

\abstract{\unboldmath Augmented large language model (LLM) systems have reported high benchmark accuracies in rare disease diagnosis, yet their real-world clinical utility remains unproven. A recent meta-analysis flagged augmented LLM diagnostic evaluations as having a high risk of bias due to suspected data leakage; however, the precise impact of these vulnerabilities on model inference has yet to be fully quantified. In this study, we present a comprehensive empirical audit to quantify evaluation artifacts across prevailing AI paradigms, using a repository of over 60,000 cases from diverse public and independent clinical cohorts. By reconstructing a state-of-the-art agentic framework, we demonstrate that candidate-space contamination artificially inflates performance: enforcing strict cross-dataset isolation resulted in a marked performance decline, with the cross-dataset mean Recall@1 decreasing from 54.3\% (under conventional one-vs-rest evaluation) to 40.3\% (Hedges' $g=0.94$). Similar spatial overfitting in supervised classifiers caused diagnostic accuracy to drop from an internal mean of 42.9\% to 16.7\% across external cohorts (Cohen's $d=2.01$). Furthermore, we quantify how LLMs infer diagnoses by exploiting retrospective therapeutic cues in clinical narratives, and how temporal misalignment in foundational knowledge bases leads to performance fluctuations. We conclude that prevailing evaluation protocols for augmented LLM systems inadvertently reflect data leakage rather than genuine medical reasoning. To address these systematic artifacts, we propose a rigorous evaluation framework---coupling strict data-isolation protocols with standardized benchmarking safeguards---to ensure future diagnostic AI for rare diseases is evaluated on actual clinical robustness.\vspace{-0em}}

\maketitle


\section{Introduction}
Rare diseases (RDs) collectively affect an estimated 300 million people worldwide\cite{the_lancet_global_health_landscape_2024}. More than 7,000 RDs have been documented, with nearly 80\% having a genetic aetiology\cite{haendel_how_2020}. Owing to the rarity and phenotypic heterogeneity of individual conditions\cite{wakap_estimating_2020}, patients typically endure a protracted ``diagnostic odyssey'' characterized by years of redundant investigations and repeated misdiagnoses\cite{faye_time_2024,bauskis_diagnostic_2022,michaels-igbokwe_standardized_2021}. In recent years, phenotype-based artificial intelligence (AI) for diagnostic decision support has emerged as a promising solution to bridge this diagnostic gap. The field has rapidly progressed from early knowledge-driven symbolic reasoning\cite{kohler_clinical_2009,bauer_bayesian_2012,robinson_interpretable_2020,yang_robust_2023} to deep-learning pattern recognition\cite{mao_phenotype-based_2025,wen_phenodp_2025}, and more recently, to generative large language models (LLMs)\cite{vaswani_attention_2017}, domain-specific fine-tuned LLMs\cite{yang_specialized_2025,schumacher_rare_2025,wang_multimodal_2026}, and agentic systems\cite{yang_rdguru_2025,chen_enhancing_2025,chen_rareagents_2026,zhao_agentic_2026}. Accompanying this architectural evolution, recent studies have reported high diagnostic accuracies relying solely on phenotypic features or clinical narratives. These systems frequently report top-1 accuracies exceeding 50\%\cite{zhao_agentic_2026}, and sometimes nearly 70\%\cite{yang_specialized_2025}, across public benchmarks and internal clinical cohorts. Despite these advancements, such benchmark performance frequently fails to translate into real-world clinical utility, raising questions about the actual generalizability of these systems\cite{goldenberg_is_2026}.

A recent systematic review and meta-analysis evaluating the diagnostic accuracy of LLM-based systems for RD diagnosis highlighted substantial vulnerabilities within the current evidence base\cite{nguyen_diagnostic_2026}. While reporting seemingly high pooled diagnostic accuracies, the meta-analysis simultaneously revealed extreme performance heterogeneity across LLM-based systems. Notably, nearly all of the augmented LLM systems---incorporating retrieval, agent-based reasoning, or fine-tuning---were rated at a high risk of bias. This was primarily due to suspected data leakage, including training corpus overlap, retrieval database contamination, and the use of public benchmark cases that pre-date LLM training cutoffs. However, because these vulnerabilities were identified solely through retrospective analyses of metadata and methodological descriptions from published studies, the precise mechanistic impact of these factors on actual model inference has yet to be experimentally quantified. It remains unclear how much of the reported state-of-the-art (SOTA) performance stems from genuine medical reasoning capabilities versus the unintended exploitation of flawed evaluation protocols\cite{ghassemi_false_2021}.

Motivated by this critical gap, we hypothesized that prevailing architectural paradigms in modern RD diagnosis---especially augmented LLM systems---are particularly vulnerable to specific forms of structural data leakage. To systematically audit the claims of the prior meta-analysis, we focused on four dimensions likely to introduce evaluation artifacts:
First, current patient-based retrieval and retrieval-augmented generation (RAG) systems are highly susceptible to \textit{candidate-space contamination}. These systems frequently retrieve reference cases that share the same publication source, familial pedigree, or cohort origin as the query, thereby matching based on data provenance rather than true phenotypic similarity\cite{cheng_survey_2025}.
Second, domain-specific fine-tuned LLMs as well as supervised classifiers trained and tested within a single clinical cohort often exploit \textit{spatial overfitting}---memorizing institutional diagnosis-phenotype profiles, disease prevalence, and recording preferences rather than learning generalizable phenotype-disease associations\cite{kapoor_leakage_2023}.
Third, when evaluating zero-shot LLMs on clinical narratives, models can exploit retrospective \textit{post-hoc cue leakage}---such as targeted therapeutic interventions and downstream laboratory results recorded in discharge summaries---to infer the correct diagnosis\cite{cho_building_2026}.
Finally, \textit{temporal knowledge bias} arises not only when historical cases are evaluated against contemporary knowledge snapshots without chronological synchronization (inadvertently introducing future-data leakage), but also when competing models rely on disparate temporal versions of the underlying knowledge, creating an asymmetric and unfair baseline comparison.

To empirically address these risk-of-bias concerns, we present a comprehensive reanalysis and quantitative audit designed to systematically stress-test phenotype-based RD diagnostic capabilities. Using a multi-tiered repository of over 60,000 RD cases---spanning eight public literature/registry datasets and two independent clinical cohorts---we reconstructed and evaluated the specific environments of classical knowledge-driven computational models, supervised classifiers, and contemporary LLM paradigms. Specifically, we replicated the current SOTA agentic system DeepRare\cite{zhao_agentic_2026} to audit its diagnostic capability across controlled data isolation regimes while additionally validating across alternative foundation LLM backbones to confirm our findings are not model-specific. In parallel, we executed an extensive series of cross-paradigm experiments to isolate and quantify all suspected sources of evaluation artifacts across diverse diagnostic architectures. By systematically applying controlled ablation setups---enforcing cross-institutional isolation, PMID-level deduplication, clinical narrative redaction, and chronological synchronization---we quantitatively dissect how AI models inadvertently exploit structural biases to artificially inflate diagnostic metrics.

Similar to historical methodological reforms in clinical medicine---such as blinding, randomization, and protocol pre-registration\cite{moher_consort_2010}---we argue that diagnostic AI for RDs requires a similar structural shift in its benchmarking standards, aligning more closely with guidelines like CONSORT-AI\cite{liu_reporting_2020} and STARD-AI\cite{sounderajah_stard-ai_2025}. Synthesizing our systematic audit, we propose a rigorous evaluation framework. By mitigating data leakage while incorporating standardized benchmarking safeguards, this framework seeks to ensure that future diagnostic systems for RDs are judged not by their susceptibility to evaluation biases, but by their genuine clinical safety, robustness, and real-world reproducibility.

\section{Results}

\subsection{Evaluation framework}
Augmented LLM systems have demonstrated strong diagnostic performance in RDs, largely by integrating external biomedical knowledge bases (KBs), historical case retrieval, as well as domain-specific fine-tuning. Disentangling generalizable clinical reasoning from potential evaluation artifacts is essential for establishing rigorous validation standards in this domain. To examine these dynamics systematically, we designed a multi-paradigm evaluation framework. Specifically, using DeepRare---a representative agentic system---as a primary case study, we audited its reference-case retrieval mechanisms across controlled candidate bank partitions to quantify potential data overlap. To provide a comprehensive assessment across diagnostic paradigms, our framework further incorporates traditional supervised classifiers to assess cohort-level spatial overfitting, and classical knowledge-driven computational models to evaluate temporal knowledge bias.

To operationalize this audit across varying levels of annotation density, semantic noise, and potential candidate overlap, we compiled a multi-source evaluation corpus comprising over 60,000 RD cases. We structured these datasets into a three-tiered hierarchy representing diverse data origins and curation practices, enabling the systematic evaluation of artifacts and data-leakage pathways (Table~\ref{tab:dataset_summary}):

\begin{itemize}[leftmargin=*, labelsep=0.5em]
    \item \textbf{Literature-derived cohorts}: Datasets curated from published biomedical case reports, including the GA4GH Phenopacket Store\cite{danis_corpus_2025}, LIRICAL\cite{robinson_interpretable_2020}, PhenoDP\cite{wen_phenodp_2025}, RDmaster\cite{yang_rdmaster_2024}, MME\cite{philippakis_matchmaker_2015}, and RareArena\cite{chen_rarearena_2026}, alongside two published clinical literature cohorts (HMS\cite{ronicke_can_2019} and VPI\cite{stark_clinically_2017}). These cases feature expert-curated Human Phenotype Ontology (HPO)\cite{gargano_human_2024} annotations and are structurally indexed to source publications via PubMed identifiers (PMIDs).
    \item \textbf{Public electronic registries}: Clinical records compiled from established online disease repositories, represented by the Rare Metabolic Diseases Database (RAMEDIS)\cite{topel_ramedis_2006}, serving as a benchmark for evaluating specialty-specific registry data.
    \item \textbf{Independent clinical cohorts}: Two real-world clinical datasets from tertiary academic medical centers: the Children's Hospital of Zhejiang University School of Medicine (ZJUMCH, in-house) cohort and Peking Union Medical College Hospital (PUMCH) cohort\cite{mao_phenotype-based_2025}. The ZJUMCH cohort features manually curated phenotypes, whereas the PUMCH cohort employs automated phenotypic extraction, reflecting the higher annotation density and semantic redundancy typical of routine electronic health records (EHRs). As unpublished clinical records, these strictly isolated cohorts provide an essential baseline for assessing zero-leakage generalizability and evaluating clinical narratives free from publication bias.
\end{itemize}

\begin{table}[t!]
    \caption{\textbf{Overview of benchmark datasets.}}
    \label{tab:dataset_summary}
    \vspace{-0.6em}
\begin{tabularx}{\linewidth}{
    c 
    >{\raggedright\arraybackslash}X
    >{\raggedright\arraybackslash}p{2.8cm} 
    c 
    >{\raggedleft\arraybackslash}p{1.4cm} 
    >{\raggedleft\arraybackslash}p{1.4cm} 
    >{\raggedleft\arraybackslash}X
}
    \toprule
    \textbf{No.} & \textbf{Dataset} & \textbf{Country/Region} & \textbf{Case-level PMID} & \textbf{Total Cases} & \textbf{Unique Diseases} & \textbf{Mean HPO Terms} \\
    \midrule
    \multicolumn{6}{l}{\textit{Literature-derived cohorts}} \\
    -    & RareArena    & Multi-country    & $\checkmark$ & 49,760 & 4,597 & 11.5 \\
    I    & Phenopacket  & Multi-country    & $\checkmark$ & 8,101  & 516   & 9.3  \\
    II   & LIRICAL      & Multi-country    & $\checkmark$ & 384    & 262   & 11.2 \\
    III  & PhenoDP      & Multi-country    & $\checkmark$ & 130    & 72    & 14.2 \\
    IV   & RDmaster     & Multi-country    & $\checkmark$ & 201    & 62    & 10.0 \\
    V    & MME          & Multi-country    & $\checkmark$ & 46     & 20    & 11.9 \\
    VI   & HMS          & Germany          &              & 66     & 29    & 10.9 \\
    VII  & VPI          & Australia        &              & 47     & 48    & 4.1  \\
    \addlinespace[0.4em]
    \multicolumn{6}{l}{\textit{Public electronic registries}} \\
    VIII & RAMEDIS      & Germany (mainly) &              & 802    & 92    & 7.4  \\
    \addlinespace[0.4em]
    \multicolumn{6}{l}{\textit{Independent clinical cohorts}} \\
    IX   & PUMCH        & China            &              & 1,063  & 405   & 31.2 \\
    X    & ZJUMCH       & China (in-house) &              & 896    & 138   & 9.4  \\
    \bottomrule
\end{tabularx}
    \footnotesize {Note: RareArena is for development use only.}
\end{table}

Using this multi-tiered framework, our systematic investigation examines four primary dimensions of evaluation artifacts: (i)~candidate-space contamination in similar patient retrieval, (ii)~cohort-level spatial overfitting, (iii)~post-hoc textual leakage in clinical narratives, and (iv)~temporal misalignment in foundational KBs. As detailed in the following sections, our findings illustrate how agentic systems and other diagnostic paradigms can inadvertently exploit these structural vulnerabilities, underscoring the necessity for rigorous evaluation standards before AI systems can be reliably translated into clinical practice.

\subsection{Baseline auditing on the state-of-the-art agentic system}
To assess the clinical robustness of current retrieval-augmented diagnostic systems, we used DeepRare, a recently proposed SOTA agentic system, as our representative baseline. DeepRare incorporates typical agentic workflows, including external diagnostic tool integration, similar-patient retrieval, web-based knowledge search, and multi-turn diagnostic reflection. We fully replicated the system based on its code repository, preserving all core analytical functions. To ensure our reanalysis reflects the full capabilities of DeepRare and avoids reliance on outdated LLMs, we first operationalized the DeepRare architecture using the recent foundation LLM, DeepSeek-V4-Flash. Because the exact reference-patient corpus utilized in the original DeepRare study is not fully publicly accessible, we populated the system's retrieval database using our curated cohorts, specifically focusing on those (e.g., LIRICAL, MME, HMS, and RAMEDIS) that overlap with the original publication's evaluation sets. To this end, we implemented an information content-based phenotype-similarity algorithm to drive the patient retrieval module, which we hypothesized to be the primary vector for evaluation artifacts and data leakage.

Our reanalysis evaluated the system's diagnostic stability under varying data isolation conditions. In standard evaluations of such agentic systems, including DeepRare, a one-versus-rest (OVR) strategy (or random/temporal splitting) is typically employed, where only the specific deduplicated test case is withheld from the retrieval corpus. To rigorously test for structural data leakage, we compared the OVR baseline against a strict ``Full Test'' strategy. Under the full-test condition, we enforced complete spatial and literature-level isolation: the entire dataset to which the target case belongs was removed from the retrieval corpus, and strict deduplication was applied to ensure no reference cases originating from the same literature source (identified via PMIDs) could be retrieved. This strict isolation genuinely benchmarks the system's diagnostic generalizability when encountering novel cases/cohorts in real-world clinical practice.

\begin{figure}[t!]
    \centering
    \includegraphics[width=1.0\textwidth]{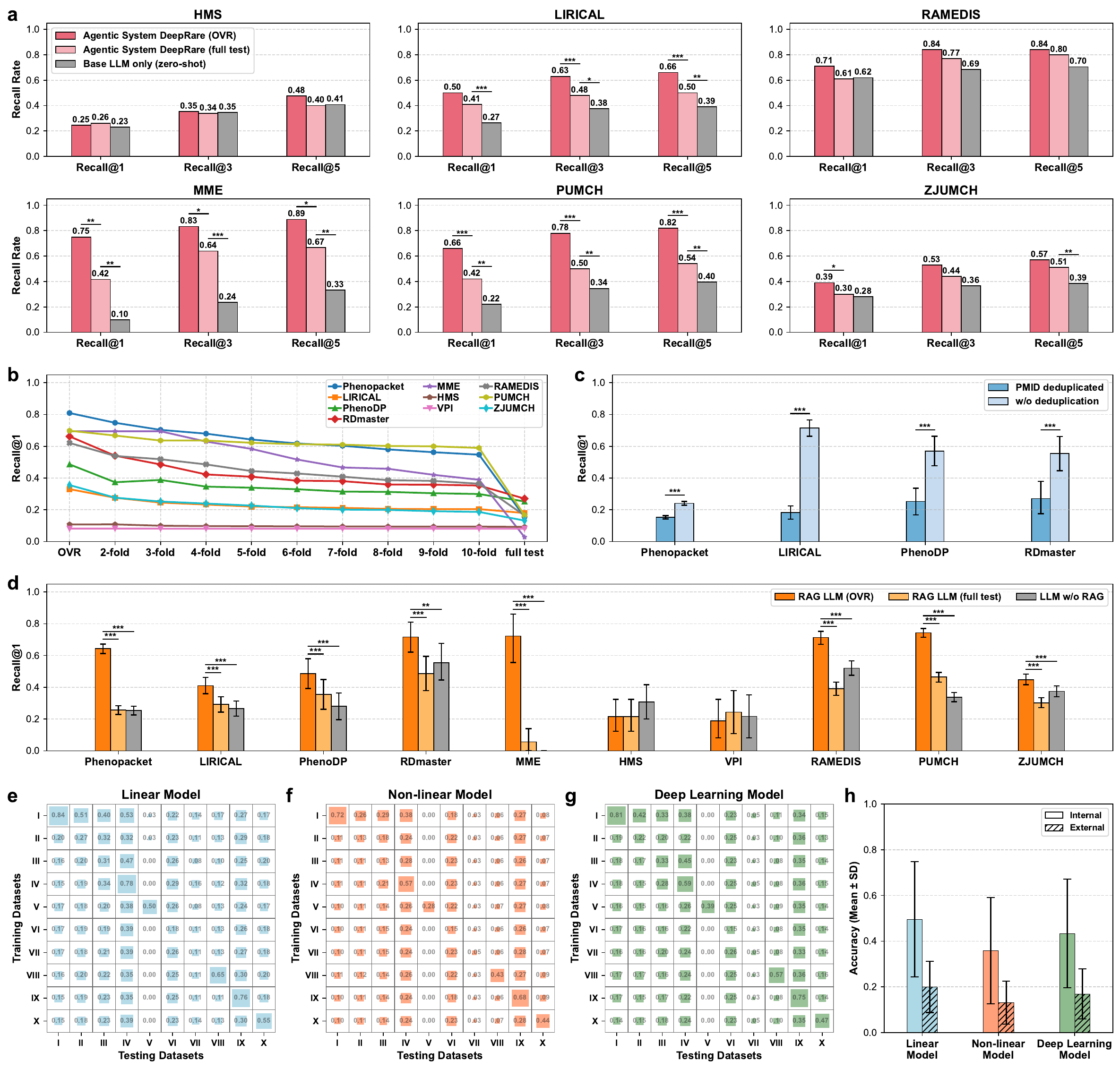} 
    \caption{\textbf{Structural data leakage and generalization collapse across rare disease diagnostic AI paradigms.}
    \textbf{(a)} Diagnostic performance of the agentic system DeepRare across six evaluation cohorts under one-vs-rest (OVR), full-test, and zero-shot LLM configurations.
    \textbf{(b)} Diagnostic Recall@1 of the standalone patient retrieval module across 10 cohorts under candidate bank partitioning schemes, transitioning from OVR through $n$-fold variants to full-test. 
    \textbf{(c)} Diagnostic Recall@1 (Mean $\pm$ 95\% CI) across four literature-derived cohorts evaluated under the full-test setting with and without cross-dataset PMID-level deduplication.
    \textbf{(d)} Diagnostic Recall@1 (Mean $\pm$ 95\% CI) of the minimalist RAG prototype across ten datasets evaluated under OVR, full-test, and standalone LLM without RAG configurations.
    \textbf{(e--g)} Pairwise cross-dataset diagnostic Recall@1 for (e) linear (SGDClassifier), (f) non-linear (XGBoost), and (g) deep learning (multilayer perceptron) models (datasets I--X correspond to Table~\ref{tab:dataset_summary}; the internal numbers indicate Recall@1).
    \textbf{(h)} Summary comparison of internal versus external validation accuracy (Mean $\pm$ SD) across supervised linear, non-linear, and deep learning classifiers.
    Note: *$P < 0.05$, **$P < 0.01$, ***$P < 0.001$.} 
    \label{fig:deeprare}
\end{figure}

The empirical results reveal a marked decline in generalization when strict isolation is enforced (Fig.~\ref{fig:deeprare}a). Under conventional OVR conditions, our replicated agent achieved benchmark metrics consistent with those reported in the original publication (e.g., Recall@1 $>70\%$ on MME and RAMEDIS, and $\geq 50\%$ on LIRICAL), confirming the successful deployment and functional fidelity of the underlying reasoning pipeline. However, when subjected to the full-test isolation, diagnostic performance declined across cohorts: the cross-dataset mean Recall@1 decreased from $54.3\% \pm 18.2\%$ under OVR to $40.3\% \pm 11.1\%$ under the full-test setting. At the individual dataset level, Recall@1 on MME dropped from 75\% down to 42\%, while the clinical cohorts from PUMCH and ZJUMCH experienced reductions from 66\% to 42\%, and from 39\% to 30\%, respectively (all $P<0.05$, paired McNemar's test). To quantify the magnitude of this artifact beyond individual cohort variations, we calculated the standardized effect sizes using Hedges' $g$ across all six evaluation cohorts. Although the agentic architecture yielded genuine gains over the zero-shot base LLM under full-test isolation (Hedges' $g=[0.77, 0.81, 1.01]$ for Recall@$[1, 3, 5]$, respectively), the artificial performance inflation induced by transitioning from strict full-test isolation to the conventional OVR setup was substantially larger (Hedges' $g=[0.94, 1.18, 1.21]$ for Recall@$[1, 3, 5]$). This comparative effect size analysis yields a critical insight: while the multi-step agentic framework does provide a measurable diagnostic advantage over raw foundation LLMs under full-test isolation, the artificial performance inflation imparted purely by candidate-space leakage (OVR artifact) is consistently larger than the system's actual reasoning contribution.

To ensure these findings were not specific to a single foundation LLM, we evaluated the DeepRare agentic architecture using diverse foundation LLMs, including Gemini-2.5-Flash and GPT-4o-mini (Supplementary Fig. S1). While the absolute baseline accuracies naturally fluctuated depending on the intrinsic medical reasoning capabilities of the respective foundation LLMs, the relative performance degradation remained consistent. Across all tested LLM backbones, the transition from OVR to full-test isolation resulted in a comparable degradation in recall metrics. This cross-LLM consistency confirms that the observed performance disparity is an architectural artifact---stemming from the retrieval module's unintended access to spatially or literarily proximate data---rather than a limitation of any single foundation LLM.

\subsection{Candidate-space contamination and spatial overfitting across AI paradigms}
To elucidate the mechanistic drivers behind the observed performance degradation in the agentic system, we conducted a detailed audit of its core reference-patient retrieval module. Case-based reasoning systems inherently prioritize diagnostic candidates by matching query cases against historical clinical precedents. We hypothesized that this mechanism is highly vulnerable to structural data contamination, acting as an algorithmic shortcut when homologous cases are retained in the reference bank. To quantify this, we permuted the candidate bank partitioning across datasets, transitioning from the OVR setup to increasingly restrictive $n$-fold variants (where only $1/n$ of the cohort was retained for retrieval), culminating in the full-test setting. Across all configurations, we ensured that the retrieval bank contained no reference cases originating from the same source publication as the query case. The results demonstrated a ``few-patient surge effect'' (Fig.~\ref{fig:deeprare}b). Under the OVR configuration, the standalone retrieval module achieved high Recall@1 across the majority of datasets ($48.5\%\pm 24.3\%$). However, as the fraction of the cohort available for retrieval diminished---with $n$ scaling from 2 to 10 in the $n$-fold setups---diagnostic performance exhibited a consistent monotonic decline. Notably, even under the 10-fold setting---where only 10\% of the target cohort was accessible within the reference bank---diagnostic accuracy remained artificially elevated relative to the full-test setting (Hedges' $g=0.85$). Only under complete full-test isolation did the diagnostic Recall@1 decrease to its baseline of $15.2\%\pm 7.0\%$. This indicates that even minimal exposure to same-source cohort data allows patient retrieval modules to exploit center-specific phenotype recording preferences, local documentation habits, or extraction noise, rather than performing generalized clinical matching.

Furthermore, our findings demonstrate that even under the full-test isolation, public benchmarks derived from biomedical literature (e.g., Phenopacket, LIRICAL, PhenoDP, and RDmaster) remain vulnerable to an overlooked source of literature leakage. To evaluate the consequence of omitting literature-level controls, we compared the full-test configuration with and without cross-dataset PMID-level deduplication (Fig.~\ref{fig:deeprare}c). Across all four literature-derived cohorts, diagnostic recall prior to deduplication was significantly higher than after strict filtering ($P<0.001$, paired one-sided bootstrap test), exhibiting a large overall effect size (Hedges' $g=1.22$). Specifically, excluding reference cases sharing the query patient's exact PMID caused diagnostic recall on LIRICAL to drop from 71\% to 18\%. This indicates that without PMID-level deduplication on literature-derived datasets, benchmark performance is largely driven by data leakage from matching cases originating from the same publication or even the exact same patient. Moreover, this vulnerability poses a potential risk for agentic frameworks equipped with web or literature search capabilities (such as DeepRare), as they can directly retrieve the query patient's original source paper when tested on literature cohorts---causing reported performance to reflect search and indexing capability rather than clinical deductive reasoning.

To determine whether this candidate-space contamination is specific to complex agentic systems or present in the broader retrieval-augmented paradigm, we designed a minimalist RAG LLM prototype built upon GPT-4o. This prototype used the exact same patient-similarity retrieval module but removed the complex agentic workflows, substituting them with a few-shot similar-case inference framework. Consistent with DeepRare, this baseline RAG LLM exhibited a comparable performance degradation when transitioning from the OVR setup to the full-test setting (Fig.~\ref{fig:deeprare}d). Notably, under the OVR strategy, this minimalist RAG prototype achieved diagnostic performance closely comparable to that of the complex DeepRare agent. This indicates that the artificial performance inflation induced by OVR leakage confounds the evaluation metrics, thereby obscuring the actual contributions of advanced agentic architectures.

The evaluation artifacts induced by data leakage in retrieval-augmented and agentic systems fundamentally parallel the classical overfitting phenomenon in supervised learning and LLM fine-tuning. To examine the universality of these validation vulnerabilities, we extended our audit framework to classical supervised machine learning and deep learning classifiers (e.g., SGDClassifier, XGBoost, and the multilayer perceptron (MLP)). Using a cross-center validation matrix, we observed substantial spatial overfitting (Fig.~\ref{fig:deeprare}e--g). When trained and evaluated within the same dataset, supervised models demonstrated artificially inflated performance by exploiting center-specific diagnosis-phenotype associations. However, when evaluated across external cohorts, diagnostic generalizability decreased substantially (Fig.~\ref{fig:deeprare}h): mean accuracy fell from 49.5\% to 20.0\% for the linear SGDClassifier (Cohen's $d=2.21$), from 35.9\% to 13.2\% for the non-linear XGBoost (Cohen's $d=1.94$), and from 43.4\% to 16.9\% for the deep learning MLP (Cohen's $d=2.04$). Aggregated across all three classifiers, the overall mean accuracy declined from 42.9\% internally to 16.7\% externally (pooled Cohen's $d=2.01$). This out-of-distribution (OOD) performance degradation is equally evident in domain-specific fine-tuned LLMs. For example, in MINT\cite{wang_multimodal_2026}, diagnostic accuracy dropped from 47.6\% on diseases seen during fine-tuning to only 7.0\% on completely unseen RD categories. Similarly, RareSeek reported a 29.2-percentage-point reduction in diagnostic accuracy when evaluated on the independent external benchmark RareBench\cite{yang_specialized_2025} (39.2\%) compared to its internal validation set (68.4\%)\cite{yang_specialized_2025}. This widespread OOD performance degradation demonstrates that evaluating diagnostic models within their training distributions is an inadequate evaluation protocol, as in-distribution testing merely rewards localized data memorization while obscuring true real-world failure modes.

\subsection{Post-hoc cue leakage in clinical narratives}
In addition to candidate-space contamination, post-hoc cue leakage within diagnostic inputs constitutes another major source of evaluation artifacts. Existing diagnostic architectures generally adopt one of two paradigms: structured HPO-based reasoning (utilized by traditional tools like PubCaseFinder\cite{fujiwara_pubcasefinder_2018} and Phenomizer\cite{kohler_clinical_2009}, or agentic pipelines like DeepRare that parse text into HPO terms) versus end-to-end reasoning over clinical narratives. Whereas HPO standardization inherently removes extraneous text, direct inference on clinical narratives remains vulnerable to post-hoc textual leakage, allowing models to exploit retrospective diagnostic hints rather than performing genuine phenotypic deduction. To quantify this leakage, we conducted a field-level ablation study across different sections of inpatient clinical records. We curated 796 complete inpatient EHRs with confirmed ORPHA diagnoses from the in-house ZJUMCH cohort. These structured EHRs comprise multiple standardized clinical fields, which we organized into four chronological stages: chief complaint (CC, encompassing demographic information and history of present illness), physical examination (PE), ancillary examinations (AE, containing specialized imaging and laboratory findings), and clinical treatment (CT). To eliminate direct label leakage, admission and discharge diagnosis fields were strictly excluded. Furthermore, any explicit mentions of the ground-truth diagnosis names within the remaining fields were programmatically redacted using regular expression matching. In parallel, standardized HPO terms were extracted from the records to serve as an isolated phenotypic representation free from narrative context. Finally, we supplied these segmented EHR sections (added chronologically) as well as the structured HPO terms to a zero-shot prompted GPT-4o model for RD differential diagnosis.

\begin{figure}[t]
    \centering
    \includegraphics[width=0.9\textwidth]{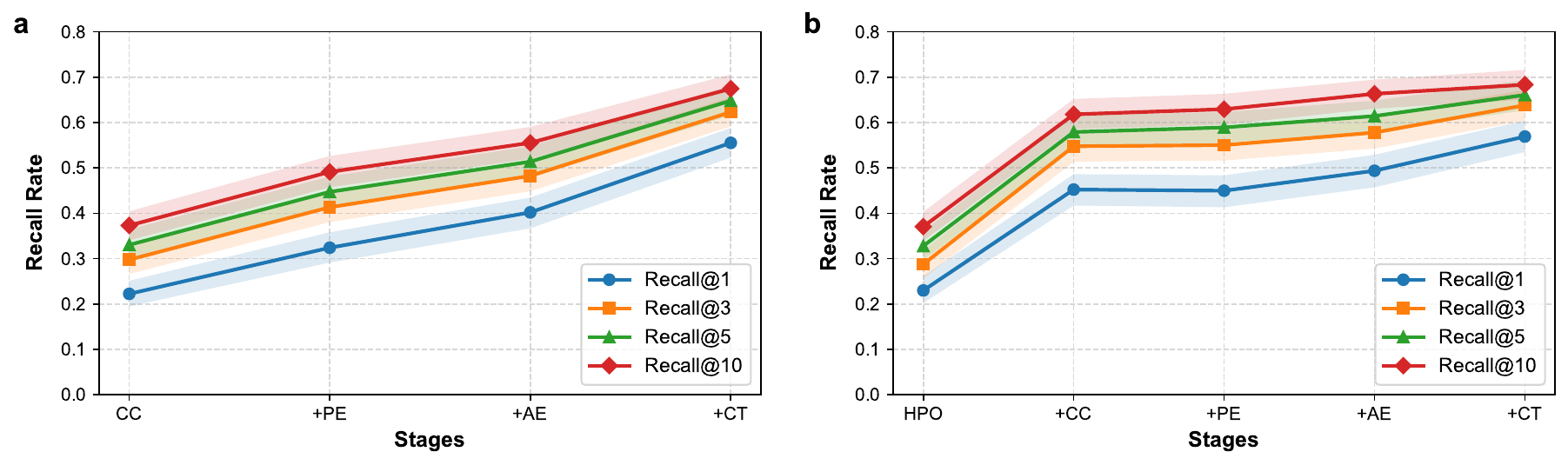} 
    \caption{\textbf{Post-hoc cue leakage in clinical narratives.} \textbf{(a)} Stepwise performance inflation (Mean $\pm$ 95\% CI) of zero-shot LLM diagnostics as retrospective clinical narrative segments---chief complaint (CC), physical examination (PE), ancillary examinations (AE), and clinical treatment (CT)---are sequentially appended. \textbf{(b)} Diagnostic recall (Mean $\pm$ 95\% CI) with sequential addition of clinical text segments to extracted HPO terms.
    }
    \label{fig:posthoc}
\end{figure}

As shown in Fig.~\ref{fig:posthoc}a, while adding phenotypic descriptions steadily improved diagnostic performance (+PE vs. CC: $\Delta \text{Recall@1}=+10.18\%$, $\text{paired McNemar's Odds Ratio (OR)}=6.06$; +AE vs. +PE: $\Delta \text{Recall@1}=+7.79\%$, $\text{OR}=3.14$), the largest performance increase occurred when clinical treatment records were appended (+CT vs. +AE: $\Delta \text{Recall@1}=+15.33\%$, $\text{OR}=6.81$). From a clinical workflow perspective, diagnosing a patient using a discharge summary that includes treatment information is inconsistent with clinical practice. Real-world diagnostic reasoning is a prospective process that occurs before targeted treatments are administered. The CT sections frequently contain retrospective post-hoc cues, such as highly specific targeted therapies (e.g., enzyme replacement therapies) or diagnostic drug-monitoring laboratory results. Similarly, AE sections sometimes include disease-specific ancillary findings---such as characteristic imaging features, histopathological markers, or metabolic panels---that act as diagnostic cues retrospectively. Fig.~\ref{fig:posthoc}b further isolates this artifact: appending raw contextual text (+CC, including demographics and medical history) to structured HPO terms provides situational context that yields a substantial improvement in diagnostic recall (+CC vs. HPO: $\Delta \text{Recall@1}=+22.24\%$, $\text{OR}=12.80$). In contrast, adding physical (+PE) and ancillary (+AE) examinations produces marginal changes (+PE vs. +CC: $\Delta \text{Recall@1}=-0.25\%$, $\text{OR}=0.96$; +AE vs. +PE: $\Delta \text{Recall@1}=+4.40\%$, $\text{OR}=2.59$), indicating they offer little additional phenotypic value beyond the extracted HPO terms. However, appending clinical treatment (+CT) information---which similarly lacks new phenotypic data---still generates a notable performance gain (+CT vs. +AE: $\Delta \text{Recall@1}=+7.54\%$, $\text{OR}=2.54$). This contrast confirms that the LLM derives substantial diagnostic signals from situational and therapeutic cues rather than genuine phenotypic data enrichment.

These findings highlight a critical dichotomy in evaluating AI diagnostic tools. While systems operating strictly on structured HPO representations are structurally insulated from post-hoc cue leakage, systems evaluated on raw, unstructured clinical text are particularly susceptible. For studies assessing clinical narratives to be clinically valid, evaluation protocols must mandate the redaction of therapeutic interventions, discharge outcomes, and post-diagnostic administrative markers, ensuring the systems are tested on genuinely prospective clinical presentations.
 
\subsection{Temporal misalignment in foundational knowledge bases}
Most augmented LLM systems fundamentally rely on foundational biomedical KBs (e.g., ORPHA\cite{weinreich_orphanet_2008} and OMIM\cite{hamosh_online_2021})---whether by directly querying KBs via RAG pipelines, indirectly utilizing them through external recommendation tools, or fine-tuning models on knowledge-base data. However, unlike traditional AI benchmarks that emphasize rigorous KB consistency across compared baselines, evaluations of contemporary LLM-based systems often neglect the synchronization of underlying KB versions. A prevailing assumption in diagnostic AI is that reference KB versioning fundamentally shapes benchmark metrics, with newer snapshots presumed to yield superior accuracy. Directly quantifying this temporal effect within complex augmented LLMs is challenging in practice, and is further methodologically confounded by their opaque internal parametric memory and stochastic inference. To cleanly isolate the impact of KB evolution on diagnostic performance, we deployed purely knowledge-driven architectures as objective analytical baselines for a longitudinal temporal audit. Specifically, we evaluated two representative knowledge-driven computational architectures---a likelihood ratio (LR) model\cite{robinson_interpretable_2020} and a probabilistic graphical (PG) model\cite{bauer_bayesian_2012}---across six annual snapshots of ORPHA and OMIM spanning 2021 to 2026. The diagnostic reasoning of these deterministic models is transparently derived from the underlying ontology and computational logic, ensuring that any observed performance changes reflect the true dynamics of the knowledge graph.

\begin{figure}[t!]
    \centering
    \includegraphics[width=1.0\textwidth]{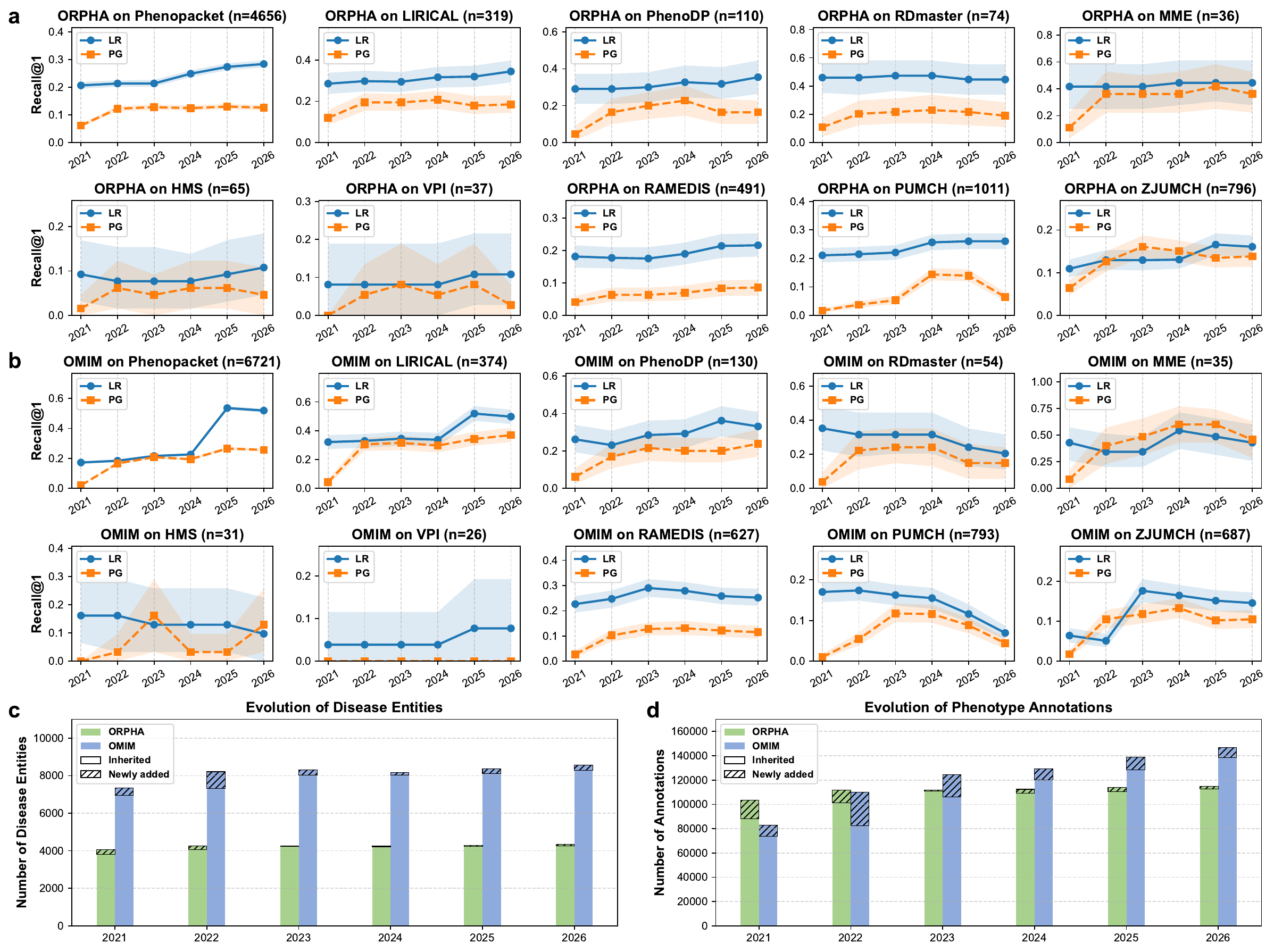} 
    \caption{\textbf{Longitudinal assessment of clinical diagnostic performance and knowledge base evolution.} \textbf{(a)} Recall@1 performance trajectories (Mean $\pm$ 95\% CI) of the likelihood ratio (LR) and probabilistic graphical (PG) models across six annual snapshots (2021--2026) of ORPHA on diverse public and clinical datasets. \textbf{(b)} Corresponding temporal evaluation using OMIM. \textbf{(c)} Temporal evolution of disease entities in OMIM and ORPHA. \textbf{(d)} Temporal evolution of disease-phenotype annotations in OMIM and ORPHA.
    }
    \label{fig:kb}
\end{figure}

Evaluating diagnostic baselines across longitudinal snapshots of foundational KBs revealed substantial performance variations tied directly to database versioning (Fig.~\ref{fig:kb}). On literature-derived datasets, models exhibited marked step-like shifts corresponding to major curator re-annotation events; for instance, the LR model's Recall@1 on Phenopacket rose from 17.2\%--22.6\% (2021--2024) to 53.6\% in the 2025 OMIM snapshot, with a parallel increase from 33.7\% to 51.9\% observed on LIRICAL (Fig.~\ref{fig:kb}b). This observation demonstrates that a model's apparent diagnostic capability on literature-derived datasets can increase substantially solely due to KB evolution, independent of algorithmic advancements. Conversely, on real-world clinical cohorts, KB evolution led to non-monotonic, bidirectional variations---such as LR Recall@1 on PUMCH fluctuating from 17.0\% in 2021 to 6.9\% in 2026---as progressive disease refinement in KBs can introduce semantic mismatches when matched against redundant clinical documentation. Moreover, different computational models exhibited divergent trajectories on the identical KB snapshots---for instance, in the OMIM benchmark on the HMS dataset, LR and PG models demonstrated contradictory trends---indicating that structural ontology updates impact distinct reasoning architectures unequally.

These findings challenge the assumption that newer KB versions consistently enhance diagnostic accuracy. Rather, our results demonstrate that version disparities in the underlying KB induce substantial performance fluctuations for the exact same model. When comparative benchmarks rely on disparate or unaligned KB snapshots, observed performance gaps may merely reflect ontology version discrepancies rather than genuine model superiority. Consequently, standardizing a unified KB version across all benchmarked baselines is an essential requirement for fair, controlled, and rigorous diagnostic AI evaluation.

\subsection{Auditing data leakage risk in rare disease diagnostic AI}
A recent meta-analysis evaluating LLM-based systems in RD diagnosis highlighted widespread risks of bias, identifying potential data leakage and over-optimistic performance in previous publications\cite{nguyen_diagnostic_2026}. However, this prior study primarily inferred these vulnerabilities from literature without conducting direct experimental re-analyses or quantitatively characterizing their specific impacts. Our systematic audit provides empirical evidence to substantiate these concerns. By isolating each potential evaluation shortcut through controlled ablation experiments, we demonstrate that benchmark inflations are not random variations, but measurable distortions driven by specific structural flaws: candidate-space and literature contamination (CC), spatial overfitting (SO), post-hoc cue leakage (PCL), and temporal knowledge bias (TKB). Applying these audit criteria, we systematically reassessed and re-annotated the majority of recent LLM-based systems alongside historical landmark studies, based on the methodological details reported in their original publications. To ensure annotation rigor, the initial risk assessment was conducted by author J.Y. and subsequently verified by an independent co-author, with any discrepancies resolved through consensus. The explicit grading rubric, along with the extracted methodological details and specific page numbers from the source publications supporting each judgment, are provided in Supplementary Table S1.

\begin{table}[t!]
    \caption{\textbf{Evaluation artifact audit across rare disease diagnostic AI systems.}}
    \footnotesize
\begin{tabularx}{\linewidth}{
    >{\raggedright\arraybackslash}p{3.1cm}
    >{\raggedright\arraybackslash}p{2.5cm}
    >{\raggedright\arraybackslash}X
    >{\centering\arraybackslash}p{0.6cm}
    >{\centering\arraybackslash}p{0.6cm}
    >{\centering\arraybackslash}p{0.6cm}
    >{\centering\arraybackslash}p{0.6cm}
    >{\centering\arraybackslash}p{0.6cm}
    >{\centering\arraybackslash}p{1.0cm}
}
\toprule
\textbf{System (year)} & \textbf{AI Paradigm} & \textbf{Diagnostic Features} & \textbf{Scope} & \textbf{CC} & \textbf{SO} & \textbf{PCL} & \textbf{TKB} & \textbf{Overall Risk} \\
\midrule
\text{DeepRare} (2026)\cite{zhao_agentic_2026} & Agent & HPO Terms\textsuperscript{\#} \& WES & P & + & -- & -- & + & + \\
\addlinespace[0.1em]
\text{RareAgents} (2026)\cite{chen_rareagents_2026} & Agent & HPO Terms \& Symptoms & P & + & -- & -- & + & + \\
\addlinespace[0.1em]
\text{MINT} (2026)\cite{wang_multimodal_2026} & FT LLM & HPO Terms \& Demographics & P & -- & + & -- & + & + \\
\addlinespace[0.1em]
\text{RareSeek} (2025)\cite{yang_specialized_2025} & RAG \& FT LLM & Clinical Text & P & -- & + & + & + & + \\
\addlinespace[0.1em]
\text{Deep-DxSearch} (2025)\cite{zheng_end--end_2025} & Agent \& RL & Clinical Text & P & + & + & -- & + & + \\
\addlinespace[0.1em]
\text{MEDDxAgent} (2025)\cite{rose_meddxagent_2025} & Agent & HPO Terms & R & + & -- & -- & + & + \\
\addlinespace[0.1em]
\text{RareScale} (2025)\cite{schumacher_rare_2025} & RAG \& FT LLM & Clinical Text & R & + & + & -- & ? & + \\
\addlinespace[0.1em]
\text{MAC} (2025)\cite{chen_enhancing_2025} & Agent & Clinical Text & P & -- & -- & + & -- & + \\
\addlinespace[0.1em]
\text{RDguru} (2025)\cite{yang_rdguru_2025} & Agent & HPO Terms\textsuperscript{\#} & P & -- & -- & -- & ? & -- \\
\addlinespace[0.1em]
\text{RareBench} (2024)\cite{chen_rarebench_2024} & RAG LLM & HPO Terms\textsuperscript{\#} \& Clinical Text & P & + & -- & + & ? & + \\
\addlinespace[0.1em]
\text{PhenoBrain} (2025)\cite{mao_phenotype-based_2025} & Ensemble & HPO Terms\textsuperscript{\#} & P & -- & -- & -- & -- & -- \\
\addlinespace[0.1em]
\text{PhenoDP} (2025)\cite{wen_phenodp_2025} & Ensemble & HPO Terms & P & -- & -- & -- & -- & -- \\
\addlinespace[0.1em]
\text{PheLR} (2023)\cite{yang_robust_2023} & Ontology-driven & HPO Terms \& WES & P & -- & -- & -- & -- & -- \\
\addlinespace[0.1em]
\text{LIRICAL} (2020)\cite{robinson_interpretable_2020} & Ontology-driven & HPO Terms \& WES & P & -- & -- & -- & -- & -- \\
\addlinespace[0.1em]
\text{BOQA} (2012)\cite{bauer_bayesian_2012} & Ontology-driven & HPO Terms & P & -- & -- & -- & -- & -- \\
\addlinespace[0.1em]
\text{Phenomizer} (2009)\cite{kohler_clinical_2009} & Ontology-driven & HPO Terms & P & -- & -- & -- & -- & -- \\
\bottomrule
\end{tabularx}

    \label{tab:risk_audit_summary}
    \footnotesize {AI Paradigm: Agent $=$ agentic system; FT $=$ domain-specific fine-tuning; RAG $=$ retrieval-augmented generation; RL $=$ reinforcement learning; Ensemble $=$ multi-model ensemble; Ontology-driven $=$ ontology-driven computation.\\
    Diagnostic Features: HPO Terms\textsuperscript{\#} $=$ extract HPO terms from text; WES $=$ whole-exome sequencing.\\
    Scope: P $=$ pan-rare disease spectrum; R $=$ restricted disease subset.\\
    CC/SO/PCL/TKB/Overall Risk: + $=$ High risk; – $=$ Low risk; ? $=$ Unclear.}
\end{table}

Our audit reveals distinct, paradigm-specific patterns of evaluation risks across the surveyed systems (Table~\ref{tab:risk_audit_summary}). First, candidate-space/literature contamination is primarily present in architectures incorporating case-based retrieval modules (e.g., DeepRare, RareAgents, Deep-DxSearch, MEDDxAgent, RareScale, and RareBench), where reference databases retain cases from the same source cohort or publication without PMID-level deduplication. Second, spatial overfitting predominantly affects fine-tuned LLMs (e.g., RareSeek, MINT, RareScale) evaluated on intra-dataset splits, resulting in degraded performance when tested on external clinical cohorts or unseen disease categories. Third, post-hoc cue leakage is observed in models that directly process clinical narratives (e.g., RareSeek, MAC, RareBench), as retrospective medical records frequently contain downstream management and therapeutic details---a phenomenon corroborated in their original studies. In contrast, frameworks restricted to standardized HPO terms structurally avoid narrative post-hoc cue leakage. Fourth, temporal knowledge bias arises when systems utilizing updated KBs or live search are benchmarked against classical computational baselines without synchronizing reference database versions. The systems annotated with uncertain temporal knowledge bias largely reflect differing pre-training knowledge cutoffs across foundation LLMs. Overall, most recent LLM-based diagnostic studies exhibit risks in one or more evaluation dimensions. In contrast, classical ontology-driven methods (e.g., Phenomizer, BOQA, LIRICAL, PheLR) maintain explicit KB version control and operate on structured HPO inputs without supervised parameter tuning or patient retrieval, thereby avoiding data leakage risks. Furthermore, ensemble models integrating ontology computing with deep learning or representation learning (e.g., PhenoBrain, PhenoDP) demonstrate that deep learning pipelines can avoid these evaluation artifacts when appropriate data isolation, feature standardization, and knowledge version control protocols are enforced.

\subsection{Guidelines for trustworthy evaluation of rare disease diagnostic AI}
Based on our systematic audit and the risk-of-bias landscape of representative AI systems, we identified critical vulnerabilities that introduce evaluation artifacts into the assessment of RD diagnostic AI. To resolve these structural vulnerabilities, we propose a comprehensive benchmarking framework (Table~\ref{tab:res-tab}) that integrates our auditing results with broader consensus in translational AI evaluation\cite{omar_how_2026}. This framework establishes rigorous protocols across several dimensions: candidate-space isolation (institutional dataset segregation and PMID-level deduplication), semantically decoupled input (redaction of therapeutic and post-hoc clinical indicators from unstructured narratives), chronological synchronization (comparative models are grounded in the same version of the KBs), cross-center spatial validation (using independent external validation cohorts), as well as established safeguards and general guidelines like protocol pre-registration, multi-dataset auditing, and candidate scope harmonization.

Furthermore, as diagnostic AI systems transition from retrospective benchmarks toward regulatory evaluation, evaluation methodologies need to align with clinical trial standards like CONSORT-AI\cite{liu_reporting_2020} and STARD-AI\cite{sounderajah_stard-ai_2025}. Future clinical validations should prioritize prospective cohort designs. In such real-world workflows, AI systems process clinical data chronologically, relying on initial chief complaints, physical examinations, and baseline laboratory panels before definitive targeted therapies are initiated or recorded. Implementing this rigorous evaluation framework will provide a realistic appraisal of diagnostic performance, ensuring that digital health tools deliver verified clinical utility rather than exploiting evaluation artifacts.

\begin{table}[t!]
    \caption{\textbf{Proposed evaluation guidelines for general rare disease diagnostic AI.}}
    \vspace{-0.6em}
    \footnotesize 
\begin{tabularx}{\linewidth}{
    >{\hsize=0.6\hsize\raggedright\arraybackslash}X 
    >{\hsize=1.2\hsize\raggedright\arraybackslash}X 
    >{\hsize=1.2\hsize\raggedright\arraybackslash}X
}
    \toprule
    \textbf{Evaluation Artifact} & 
    \textbf{Manifestation} & 
    \textbf{Proposed Evaluation Guideline} \\
    \midrule
    
    \textbf{Temporal Knowledge Bias} & 
    Inconsistencies in the KB versions that models rely on lead to substantial performance fluctuations, with diagnostic accuracy shifting based on version differences. & 
    \textbf{Chronological Synchronization:} Enforce a single, chronologically aligned KB version across all models to prevent KB version-driven evaluation artifacts. \\
    \addlinespace[1em]
    
    \textbf{Spatial / Institutional Overfitting} & 
    Models memorize institutional documentation habits, cohort-specific diagnosis-phenotype associations and disease prevalence rather than learning generalizable phenotype-disease profiles. & 
    \textbf{Cross-institutional Validation:} Replace standard intra-dataset splits with mandatory out-of-distribution validation using independent, institutionally isolated clinical cohorts. \\
    \addlinespace[1em]
    
    \textbf{Candidate-space \& Literature Contamination} & 
    Retrieval systems exploit structural data leakage by matching query cases to homologous records originating from the exact same publication, family pedigree, or institution. & 
    \textbf{Source-level Deduplication:} Exclude structurally and institutionally related cases from the reference bank, enforcing PMID-level cross-deduplication for literature-derived cohorts. \\
    \addlinespace[1em]
    
    \textbf{Post-Hoc Cue Leakage} & 
    Retrospective clinical narratives embed downstream therapeutic interventions and targeted lab results, allowing LLMs to infer diagnoses retrospectively. & 
    \textbf{Semantic Decoupling \& Sanitization:} Benchmark primarily using standardized phenotypic profiles such as HPO terms. When evaluating raw clinical texts, systematically redact post-hoc therapeutic and administrative cues. \\
    \addlinespace[1em]
    
    \textbf{Post-hoc Hypothesis Shifting} & 
    Iterative prompt engineering and hyperparameter tuning directly on the evaluation set cause benchmark overfitting and compromise generalizability. & 
    \textbf{Blinded Evaluation \& Pre-registration:} Lock model weights, prompt templates, and evaluation pipelines prior to testing. Utilize independent validation sets to prevent test-set optimization. \\
    \addlinespace[1em]

    \textbf{Dataset Heterogeneity} & 
    Substantial variations in diagnostic difficulty and disease spectrum across case datasets make single-source evaluations unrepresentative. & 
    \textbf{Multi-tiered Cross-dataset Auditing:} Diagnostic systems must be benchmarked across multi-tiered datasets to reliably verify generalization across diverse data origins and clinical complexities. \\
    \addlinespace[1em]    

    \textbf{Diagnostic Scope Inconsistency} & 
    Inconsistencies in candidate diagnostic spaces across competing models introduce substantial evaluation bias. & 
    \textbf{Scope Harmonization \& Full-spectrum Coverage:} Benchmarks must align the diagnostic candidate space across all evaluated models to ensure fair comparisons. A general RD diagnostic AI should cover the broadest spectrum of RDs. \\
    
    \bottomrule
\end{tabularx}
    \label{tab:res-tab}
    \footnotesize {Abbreviations: RD, rare disease; AI, artificial intelligence; KB, knowledge base; EHR, electronic health record; PMID, PubMed identifier; LLM, large language model; HPO, Human Phenotype Ontology.}
\end{table}

\section{Discussion}
Despite advancements in phenotype-based RD diagnostics, high performance metrics reported in recent benchmarks often fail to translate into real-world clinical utility\cite{nguyen_diagnostic_2026}. Previously, the extent to which these results were driven by genuine medical reasoning versus evaluation artifacts remained unclear. In this study, we demonstrate that current AI evaluation protocols are confounded by systematic data leakage and validation biases. Through a multi-paradigm audit based on over 60,000 RD cases, we quantified how distinct algorithmic shortcuts---specifically, temporal knowledge misalignment, spatial overfitting, candidate-space contamination, and post-hoc textual cues---artificially inflate diagnostic accuracy. These empirical findings highlight the vulnerabilities embedded in current diagnostic AI benchmarking and provide a clinically aligned framework to ensure the reproducibility and generalizability of future diagnostic systems.

\textbf{The confluence of big data and LLMs exacerbates leakage risks}\\
The increasing prevalence of these evaluation artifacts is driven by shifts in how RD data is curated and utilized. Historically, the clinical scarcity of documented cases precluded the development of robust supervised classifiers as well as case-based retrieval systems. However, the recent emergence of large-scale, aggregated patient repositories (e.g., GA4GH Phenopacket Store\cite{danis_corpus_2025} and RareArena\cite{chen_rarearena_2026}) has altered this landscape, making the training of supervised classifiers both feasible and seemingly robust. This data expansion has been accelerated by LLMs, which have lowered the computational and financial costs of extracting, standardizing, and curating heterogeneous patient profiles from unstructured medical literature and EHRs\cite{zhao_large-scale_2023,chen_rarearena_2026}. Furthermore, these generative tools facilitate the continuous updating of clinical KBs\cite{manda_large_2025}. While this increased data availability represents a significant advance for RD informatics, the associated risks of data contamination have not been sufficiently recognized by the broader research community\cite{cheng_survey_2025}. When researchers apply conventional computer science evaluation protocols---such as random intra-dataset splitting---to these newly available large-scale cohorts, they inadvertently introduce the spatial overfitting and candidate-space contamination identified in our audit. The large volume of accumulated patient records, combined with frequent knowledge updates, creates conditions where temporal and structural leakages occur without being readily detected. Consequently, as the development of diagnostic models becomes more accessible, the risk of unknowingly validating them on contaminated, overlapping, or temporally misaligned data increases, necessitating a systematic recalibration of benchmarking standards.

\textbf{Re-evaluating performance claims in contemporary LLM paradigms}\\
Performing differential diagnosis across thousands of RDs is computationally complex. Historically, knowledge-driven computational models maintained transparency by explicitly documenting their logic and underlying KB versions, enabling reproducible comparisons\cite{kohler_clinical_2009,bauer_bayesian_2012,robinson_interpretable_2020,yang_robust_2023}. The recent shift toward LLMs has introduced augmented LLM systems, such as the agentic system DeepRare, which reported high capabilities (e.g., achieving an HPO-based Recall@1 of 57\%, outperforming baselines like PhenoBrain's <30\%). However, our empirical reanalysis prompts a critical re-evaluation of these claims.

The DeepRare study evaluated its patient retrieval module using a one-vs-rest (OVR) methodology across public datasets, and utilized chronological splitting for intra-institutional testing. Although the DeepRare study ablated its patient retrieval module---confirming not only that disabling case retrieval caused a substantial drop in diagnostic accuracy, but also that patient retrieval contributed far more to performance gains than its knowledge search or multi-turn reflection modules\cite{zhao_agentic_2026}---it did not critically examine the validity of its retrieval strategy. It remained ambiguous whether its reported performance gains reflected genuine clinical reasoning inherent to the patient-similarity retrieval paradigm or were artificially induced by candidate-space leakage. Our work directly resolves this ambiguity. As our audit demonstrates, OVR introduces substantial candidate-space leakage, while chronological splitting fails to resolve spatial overfitting because institutional documentation profiles and extraction noise remain structurally homogeneous over time. By replicating the DeepRare architecture and subjecting it to strict full-test isolation, we empirically confirmed that its reported performance advantages are largely artifacts of evaluation design. Our minimalist RAG prototype experiments further corroborate that this vulnerability is not merely an implementation flaw in DeepRare, but a pervasive evaluation artifact across the entire case-retrieval RAG paradigm. Furthermore, DeepRare incorporates automated online web searches, dynamically absorbing contemporary knowledge and introducing a risk of retrieving the exact published case reports of the query patients. This not only circumvents temporal version control but also creates an asymmetric comparison against chronologically frozen, offline baselines like PhenoBrain. This conclusion is reinforced by a recent audit of 300 strictly independent cases (100 each from public registries, PUMCH, and ZJUMCH)\cite{yang_fusionrare_2026}, which found DeepRare's Recall@1 (32\%) to be lower than GPT-5.5 (39\%) yet higher than PhenoBrain (27\%)---suggesting its diagnostic advantage is less pronounced than originally reported.

Analogous evaluation artifacts extend to domain-specific fine-tuned LLMs. For instance, the recently introduced RareSeek model fine-tuned an LLM on internal clinical records, reporting an in-distribution Recall@1 of 68.4\%, yet its performance dropped to 39.2\% on the independent external benchmark RareBench. Confining training and evaluation to the same institutional data distribution induces substantial spatial overfitting and OOD performance degradation, directly mirroring the empirical trends observed in our $10\times 10$ cross-cohort analysis on supervised classifiers. Furthermore, models that perform inference directly over clinical notes are highly vulnerable to post-hoc cue leakage. RareSeek utilizes comprehensive inpatient records for diagnostic prediction---encompassing specialized confirmatory test results, targeted therapeutic interventions, and disease trajectory responses. As demonstrated in our field-level narrative ablation experiments, processing these retrospective clinical summaries introduces substantial post-hoc clues that artificially inflate diagnostic accuracy. This design is inconsistent with prospective clinical workflows, where differential diagnoses must be formulated prior to initiating targeted therapies and observing longitudinal disease trajectories.

\textbf{The statistical vulnerability of multi-dataset aggregation}\\
As the RD AI field increasingly adopts multi-dataset benchmarking, the substantial disparity in cohort sizes---ranging from dozens of cases (e.g., MME and VPI) to tens of thousands (e.g., Phenopacket and RareArena)---introduces a statistical vulnerability. When aggregating diagnostic performance across multiple datasets, standard micro-averaging (treating every patient equally) allows large public registries to disproportionately influence a model's performance on smaller, specialized, or real-world clinical test sets. Conversely, unweighted macro-averaging across dataset scores can allow small, low-variance datasets with high accuracy to skew the overall performance assessment of a model. To ensure transparent reporting, performance metrics should be explicitly reported at the individual cohort level. When a global summary metric is required, studies should concurrently report both macro-averaged and micro-averaged metrics, justifying the chosen aggregation method to prevent cohort size imbalances from masking generalization gaps.

\textbf{Limitations and future directions}\\
Although our systematic audit provides a methodological framework for identifying evaluation artifacts, it has several limitations. First, while the public datasets are globally sourced, the independent clinical cohorts (ZJUMCH and PUMCH) represent tertiary care centers exclusively in China. Future validation efforts should incorporate multinational, multi-ethnic prospective EHR data to investigate whether spatial overfitting manifests differently across distinct regional healthcare systems. Second, our analysis primarily focuses on phenotype- and text-driven diagnostic reasoning. As the field advances toward multimodal AI---integrating whole-exome sequencing (WES) or medical imaging---new dimensions of data leakage, such as genomic batch effects or imaging equipment footprints, may emerge and require similar auditing. Finally, the continuous, opaque updating of proprietary LLMs makes it challenging to accurately quantify training data contamination. Consequently, our findings represent a snapshot of current evaluation vulnerabilities rather than a static metric or best practice pertaining to any single model's capability. As the deployment of AI in RD diagnostics progresses, transitioning from conventional computer science benchmarks to clinical-trial-inspired validation frameworks will shift the field's focus from leaderboard performance to clinical safety. Future studies should address these challenges by developing dynamic, community-driven benchmarking platforms that enforce temporal synchronization, cross-cohort deduplication, and cross-site isolation. Furthermore, open-source initiatives should prioritize the programmatic sanitization of clinical narratives to eliminate post-hoc therapeutic cues from test sets. By identifying the sources of artificial performance inflation, this work provides the methodological groundwork for building trustworthy, generalizable medical AI---systems capable of navigating clinical uncertainty and reliably reducing diagnostic delays for patients with rare diseases.

\section{Methods}
\subsection{Historical snapshotting of rare disease knowledge bases}
To evaluate temporal knowledge alignment, we curated historical versions of two primary rare disease (RD) knowledge bases (KBs): Online Mendelian Inheritance in Man (OMIM)\cite{hamosh_online_2021} and Orphanet (ORPHA)\cite{weinreich_orphanet_2008}.
Annual snapshots spanning 2021 to 2026 were retrieved from their official repositories. Each snapshot preserves the state of disease-phenotype associations and HPO-semantic structures at that point in time, serving as a controlled baseline to assess the influence of database updates on diagnostic performance.

\subsection{Curation of multi-source patient repositories}
We compiled a dataset of over 60,000 RD cases derived from three primary sources: (1) literature-derived cohorts (RareArena\cite{chen_rarearena_2026}, Phenopacket\cite{danis_corpus_2025}, LIRICAL\cite{robinson_interpretable_2020}, RDmaster\cite{yang_rdmaster_2024}, PhenoDP\cite{wen_phenodp_2025}, MME\cite{philippakis_matchmaker_2015}, VPI\cite{stark_clinically_2017}, and HMS\cite{ronicke_can_2019}), (2) public electronic registries (RAMEDIS\cite{topel_ramedis_2006}), and (3) independent clinical EHR cohorts (ZJUMCH and PUMCH\cite{mao_phenotype-based_2025}).

To standardize these heterogeneous data sources for computational benchmarking, we implemented a preprocessing pipeline. For the independent clinical cohorts, we initially collected 1,729 raw EHRs from ZJUMCH. Phenotypic features from ZJUMCH records were manually extracted and mapped to Human Phenotype Ontology (HPO) terms. Ground-truth clinical diagnoses were subsequently mapped to canonical OMIM/ORPHA nomenclature. ZJUMCH cases that could not be mapped to a recognized OMIM/ORPHA diagnosis or contained fewer than three initial phenotypes were excluded. For the PUMCH cohort, phenotypic features were extracted by the original researchers using an automated annotation tool. In contrast, the public literature-derived cohorts and registries were acquired with pre-curated HPO annotations and standardized diagnostic labels, bypassing the raw text extraction phase.

To ensure valid cross-temporal evaluations, all integrated cases underwent a snapshot-specific filtering process before being evaluated against any historical KB version. A case was deemed valid for a specific KB snapshot (e.g., ORPHA 2023) if it met three criteria: (i)~the ground-truth OMIM/ORPHA disease existed and had associated phenotype annotations within that specific KB snapshot; (ii)~the case retained at least two valid HPO terms present in that specific snapshot; and (iii)~the case passed intra-dataset deduplication to remove identical patient records. This filtering protocol controls for data quality and chronological alignment across temporal and spatial evaluations.

\subsection{Knowledge-driven phenotypic diagnostic baselines}
We implemented two established phenotype-based computational algorithms to process a patient HPO query set $\mathcal{Q} = \{ t_1, t_2, ..., t_{n} \}$ and prioritize candidate diseases within synchronized KB snapshots, following their original implementation specifications.

\begin{itemize}[leftmargin=*, labelsep=0.5em]
\item \textbf{Likelihood ratio model:} Based on the Naive Bayes assumption (specifically LIRICAL\cite{robinson_interpretable_2020}), this approach treats observed phenotypes as independent tests. Diagnostic evidence is aggregated by summing the log-likelihood ratios of individual phenotypes based on disease-specific frequencies and background prevalences.
\item \textbf{Probabilistic graphical model:} This approach (specifically BOQA\cite{bauer_bayesian_2012}) models the HPO taxonomy as a directed acyclic graph (DAG). It enables the joint modeling of hierarchical inheritance and conditional dependencies among HPO terms, parameterizing false-positive and false-negative observation rates to calculate the global marginal posterior probability for candidate diseases.
\end{itemize}

\subsection{Case-based retrieval diagnosis framework}
The case-based retrieval framework matches a query patient $\mathcal{Q}$ against a historical case library. The diagnostic specificity of an HPO term $t$ is quantified via information content (IC), calculated as the negative logarithm of its expected prevalence across the annotation space:
\begin{equation}
\IC(t) = -\log \left( \frac{\sum_{d \in \mathcal{D}} \Pr(t \mid d)}{|\mathcal{D}|} \right)
\end{equation}
where $\Pr(t \mid d)$ is the frequency of phenotype $t$ annotated for disease $d$, and $|\mathcal{D}|$ denotes the total number of candidate RDs.

Phenotypic similarity between $\mathcal{Q}$ and a historical reference case $\mathcal{Q}_\text{ref}$ is measured by the average IC across their shared ancestors:
\begin{equation}
\Sim(\mathcal{Q}, \mathcal{Q}_\text{ref}) = \frac{1}{2} \left[ \sum_{t \in \mathcal{Q} \cap \Anc(\mathcal{Q}_\text{ref})} \IC(t) + \sum_{t' \in \mathcal{Q}_\text{ref} \cap \Anc(\mathcal{Q})} \IC(t') \right]
\end{equation}

A max-pooling strategy aggregates case-level similarity scores to generate disease-level recommendations. The priority score for candidate disease $d$ is determined by its most phenotypically similar historical counterpart.

\subsection{Supervised machine learning and deep learning}
To perform RD diagnosis on sparse phenotype features, we implemented three representative classes of classifiers: a linear model, a non-linear tree-based ensemble, and a deep feed-forward neural network.

\begin{itemize}[leftmargin=*, labelsep=0.5em]
\item \textbf{Feature representation: }The model input was constructed by mapping query phenotypes onto an $M$-dimensional multi-hot query vector $\mathbf{x} \in \{0, 1\}^{M}$, where $M$ represents the total number of active HPO terms ($M \approx 16,000$). Directed ancestor propagation was applied to capture the hierarchical semantic structure of the HPO, setting an index to 1 if the corresponding HPO term or its descendants were present. The label space consisted of $N$ unique classes mapped to OMIM or ORPHA nomenclatures.
\item \textbf{Linear classifier:} Optimized via stochastic gradient descent (SGD)\cite{pedregosa_scikit-learn_2011} with a log loss function, regularized using an $L_2$ weight penalty (regularization coefficient $\alpha=1\times 10^{-4}$).
\item \textbf{Non-linear classifier:} An extreme gradient boosting (XGBoost)\cite{chen_xgboost_2016} ensemble with a learning rate of 0.1 and 50 boosting rounds, utilizing a column-subsampling strategy (10\% of the active feature space per tree) and a maximum tree depth of 5 to prevent overfitting.
\item \textbf{Deep learning classifier:} A multilayer perceptron (MLP)\cite{rumelhart_learning_1986} consisting of an input layer that receives the phenotypic feature vector, followed by a hidden layer with 1,024 units and ReLU activation, a dropout regularization layer, and finally a fully connected output layer for classification. The model was trained with a batch size of 32 using the AdamW optimizer with an initial learning rate of $1\times 10^{-4}$ and a cosine annealing schedule.
\item \textbf{Dataset splitting:} For cross-dataset evaluations, a 50/50 split was enforced per target dataset. Training sets were constructed by merging the remaining 50\% with RareArena (the largest dataset with the broadest disease coverage).
\end{itemize}

\subsection{Large language model-based diagnostic paradigms}
We implemented three large language model (LLM)-based diagnostic paradigms: (i)~a replication of the state-of-the-art agentic system DeepRare, (ii)~a controlled, simplified patient-retrieval LLM prototype, and (iii)~a zero-shot narrative inference framework designed for stepwise clinical text ablation.

\subsubsection{Redeployment of the DeepRare agentic system}
We replicated the DeepRare framework using its open-source repository, preserving its core multi-step reasoning architecture. The agentic pipeline integrates several key functional modules: (i)~dynamic external diagnostic tool invocation (PhenoBrain and PubCaseFinder), (ii)~live web-based biomedical knowledge search, (iii)~case-based similar-patient retrieval, and (iv)~multi-turn knowledge-enhanced reflection and candidate re-ranking. Since the original reference patient database used in the DeepRare study is not publicly available, we populated the system's patient retrieval database using our curated data hierarchy (Table~\ref{tab:dataset_summary}), specifically incorporating the public and clinical cohorts that overlap with the original study (e.g., LIRICAL, MME, HMS, and RAMEDIS). Without loss of generality, we implemented DeepRare's similar-patient retrieval functionality using our proposed IC-based case retrieval framework.

\subsubsection{Patient-retrieval RAG LLM prototype}
To evaluate the patient retrieval module independently of complex agentic workflows, we developed a simplified retrieval-augmented generation (RAG) prototype. This baseline omitted components such as multi-turn reflection and tool invocation, coupling the foundation LLM directly to the IC-based case retrieval module. For a given query $\mathcal{Q}$, the patient retrieval framework retrieved the top-10 most phenotypically similar historical cases. These exemplars were provided as few-shot contexts alongside the query case to generate a prioritized differential list. Importantly, the prompt was structured such that the retrieved cases were presented as supporting clinical references to assist differential reasoning, rather than as a closed multiple-choice candidate pool.

\subsubsection{Zero-shot diagnosis on clinical narratives}
To evaluate LLM sensitivity to text-based contextual biases, we adopted a zero-shot diagnostic paradigm utilizing clinical narrative text. We curated 796 de-identified inpatient EHRs with confirmed ORPHA diagnoses from the in-house ZJUMCH cohort. To simulate the chronological progression of real-world hospital admission workflows, we segmented each raw medical record into four sequential clinical sections: (i)~chief complaint (CC): primary onset symptoms, demographic information, and history of present illness; (ii)~physical examination (PE): objective dysmorphic features and physical signs; (iii)~ancillary examinations (AE): laboratory and radiological test findings; and (iv)~clinical treatment (CT): therapeutic interventions and pharmacological treatments, with occasional inclusion of ancillary examination findings. To eliminate direct label leakage, any fields explicitly indicating diagnostic outcomes---such as admission and discharge diagnoses---were strictly excluded, and regular expression pattern matching was applied to redact all mentions of ground-truth diagnosis names across the remaining sections. We conducted an incremental ablation protocol by sequentially appending these segments to the zero-shot prompt. The LLM was instructed to infer a prioritized differential list of candidate RDs. Additionally, to determine whether narrative text provides additional phenotypic value beyond structured ontologies, we evaluated a hybrid input setting where raw narrative segments were incrementally appended to baseline structured HPO annotations extracted from clinical narratives.

\subsubsection{Disease entity resolution}
To map LLM-generated free-text diagnoses to standardized OMIM/ORPHA codes while mitigating identifier hallucinations, we implemented a semantic entity resolution pipeline using BioLORD\cite{remy_biolord-2023_2024}, a pre-trained biomedical language model. We computed dense embeddings for both the LLM-generated diagnoses and all canonical diseases in OMIM/ORPHA, assigning the OMIM/ORPHA code with the highest cosine similarity to the LLM's raw diagnoses.

\subsection{Experimental configurations and evaluation settings}
\textbf{Diagnostic metrics:} Diagnostic efficacy was evaluated using Recall@$n$ ($n \in [1, 10]$) with an ``any-hit'' strategy for multiple causative diseases. A diagnostic session was marked successful if any ground-truth causative diagnosis ranked within the top-$n$ prioritized recommendations.

\textbf{KB version control:} For temporal evaluations (2021--2026), test cases were filtered to ensure ground-truth diagnoses were annotatable under the specific annual snapshot, with HPO profiles synchronized to the corresponding ontology version of that year. Within any single snapshot-based benchmark run, patient cohort labels were restricted to either OMIM or ORPHA. For static baseline evaluations that did not require temporal knowledge permutation, we used the 2023 ORPHA snapshot as the global reference database.

\textbf{LLM configurations:} All LLMs utilized across our experimental paradigms were deployed using their official snapshots. To ensure reproducibility and prevent performance variations introduced by generation-level optimization, we did not apply any custom hyperparameter adjustments. All inference was conducted using the default generation parameters (e.g., default temperature, probability threshold for top-$p$ sampling, and frequency penalties) without any task-specific decoding configurations.

\textbf{Case retrieval strategies:} For any diagnostic paradigms that include the case retrieval module, evaluations on a target dataset $D_{target}$ were typically conducted under three reference-bank partitioning configurations: (i)~Full test: $D_{target}$ was completely excluded from the retrieval bank; (ii)~$n$-fold variant: $1/n$ of $D_{target}$ was allocated to the retrieval bank and the remainder used for testing; and (iii)~one-vs-rest (OVR): excluding only the identical record of the query patient. Any datasets in Table~\ref{tab:dataset_summary} other than $D_{target}$ were included in the retrieval bank. Additionally, because many LIRICAL cases are subsets of the Phenopacket dataset, Phenopacket cases were explicitly excluded from the retrieval bank when evaluating DeepRare on the LIRICAL cohort.

\textbf{PMID-level deduplication:} PMID-level deduplication was implemented in three steps: (i)~identifying the source PMID associated with each literature-derived query case, (ii)~matching this identifier against the publication metadata across the entire reference corpus, and (iii)~excluding all co-published reference cases from the candidate bank prior to inference.

\textbf{Supervised classifier implementation:} The linear classifier was implemented using SGDClassifier (scikit-learn v1.0.2)\cite{pedregosa_scikit-learn_2011}, the non-linear ensemble using XGBoost (xgboost v2.0.3)\cite{chen_xgboost_2016}, and the deep feed-forward neural network was implemented using PyTorch (v2.1.2)\cite{paszke_pytorch_2019}.

\textbf{Statistical analysis:}
(i)~For paired binary diagnostic outcomes across DeepRare evaluation settings, statistical significance was evaluated using McNemar's test. (ii)~A paired one-sided bootstrap test was used to assess the statistical significance of performance differences between methods as well as the 95\% confidence intervals (CIs). For each metric (e.g., Recall@$k$), we generated 2,000 bootstrap samples by sampling test cases with replacement. The one-sided $P$-value was calculated as the fraction of bootstrap iterations where each model achieved an equal or higher recall than a designated reference model, following the methodology described by Koehn\cite{koehn_statistical_2004}. The 95\% CIs were derived using the 2.5th and 97.5th percentiles of the bootstrap distribution. (iii)~The paired McNemar's odds ratio was employed to quantify relative diagnostic gains across clinical narrative ablation stages. (iv)~To quantify effect sizes across multi-dataset aggregations, we calculated Hedges' $g$ with small-sample bias correction. (v)~Cohen's $d$ was calculated to measure the performance degradation between in-distribution and out-of-distribution evaluations in supervised classifiers.

\backmatter

\section*{Data Availability}
Historical versions of the Human Phenotype Ontology and corresponding OMIM/ORPHA disease annotations are publicly accessible at \url{https://github.com/obophenotype/human-phenotype-ontology}. The multi-source rare disease patient cohorts utilized in this study are available via their respective URLs, as detailed in Supplementary Table S2. The primary electronic health record data from the Children's Hospital of Zhejiang University School of Medicine (ZJUMCH) are not publicly available due to institutional data privacy regulations and ethical constraints. The ZJUMCH patient data analyzed in this work were derived from a prior research study conducted under institutional review board (IRB) approval (ZJUMCH IRB \#2021-IRB-165, \#2022-IRB-068), and all clinical records were fully de-identified prior to computational analysis in the present study.

\section*{Code Availability}
The official source code for the DeepRare agentic framework is available at \url{https://github.com/MAGIC-AI4Med/DeepRare}. Source code for the established baseline models, LIRICAL and BOQA, was obtained from \url{https://github.com/TheJacksonLaboratory/LIRICAL} and \url{https://github.com/iamkoehler/boqa}, respectively.
All original code developed for this audit study---including the minimalist RAG baseline, the supervised classifier pipelines, and the statistical evaluation scripts---has been deposited on GitHub and is publicly available at \url{https://github.com/1678492347/RD-ReEval}.

\section*{Acknowledgements}

This work is supported by the National Natural Science Foundation of China (Nos. 62506329 and 82572377) and the ``Pioneer'' and ``Leading Goose'' R\&D Program of Zhejiang (No. 2026C01022).

\section*{Author Contributions}
J.Y. and H.L. conceived and designed the study, developed the methodology, managed project administration, and acquired funding. J.Y., P.L., X.L., and H.L. curated and processed the benchmark datasets. Y.C. and Y.F. conducted the investigations. J.Y. and P.L. performed the formal data analysis. J.Y., X.L., Y.C., Y.F., and H.L. performed the validation of results. H.L. provided resources and overall supervision. J.Y. and H.L. drafted the initial manuscript. All authors (J.Y., P.L., X.L., S.L., Y.C., Y.F., and H.L.) critically reviewed and edited the manuscript, and approved the final version.

\section*{Competing Interests}

The authors declare no competing interests.

\clearpage
\bibliography{re-eval}

@article{kohler_clinical_2009,
	title = {Clinical {Diagnostics} in {Human} {Genetics} with {Semantic} {Similarity} {Searches} in {Ontologies}},
	volume = {85},
	issn = {00029297},
	doi = {10.1016/j.ajhg.2009.09.003},
	i_language = {en},
	number = {4},
	journal = {American Journal of Human Genetics},
	author = {Köhler, Sebastian and Schulz, Marcel H. and Krawitz, Peter and Bauer, Sebastian and Dölken, Sandra and Ott, Claus E. and Mundlos, Christine and Horn, Denise and Mundlos, Stefan and Robinson, Peter N.},
	month = oct,
	year = {2009},
	pages = {457--464},
}

@article{bauer_bayesian_2012,
	title = {Bayesian ontology querying for accurate and noise-tolerant semantic searches},
	volume = {28},
	issn = {13674803},
	doi = {10.1093/bioinformatics/bts471},
	i_language = {en},
	number = {19},
	journal = {Bioinformatics},
	author = {Bauer, Sebastian and Köhler, Sebastian and Schulz, Marcel H. and Robinson, Peter N.},
	month = oct,
	year = {2012},
	pages = {2502--2508},
}

@article{weinreich_orphanet_2008,
	title = {Orphanet: a {European} database for rare diseases},
	volume = {152},
	issn = {0028-2162},
	i_url = {http://europepmc.org/abstract/MED/18389888},
	i_language = {en},
	number = {9},
	journal = {Nederlands tijdschrift voor geneeskunde},
	author = {Weinreich, S. S. and Mangon, R. and Sikkens, J. J. and Teeuw, M. E. en and Cornel, M. C.},
	year = {2008},
	pages = {518--519},
}

@article{hamosh_online_2021,
	title = {Online {Mendelian} {Inheritance} in {Man} ({OMIM}®): {Victor} {McKusick}'s magnum opus},
	volume = {185},
	i_url = {https://onlinelibrary.wiley.com/doi/abs/10.1002/ajmg.a.62407},
	doi = {https://doi.org/10.1002/ajmg.a.62407},
	i_language = {en},
	number = {11},
	journal = {American Journal of Medical Genetics Part A},
	author = {Hamosh, Ada and Amberger, Joanna S. and Bocchini, Carol and Scott, Alan F. and Rasmussen, Sonja A.},
	year = {2021},
	i_note = {\_eprint: https://onlinelibrary.wiley.com/doi/pdf/10.1002/ajmg.a.62407},
	pages = {3259--3265},
}

@article{bauskis_diagnostic_2022,
	title = {The diagnostic odyssey: insights from parents of children living with an undiagnosed condition},
	volume = {17},
	issn = {1750-1172},
	i_url = {https://doi.org/10.1186/s13023-022-02358-x},
	doi = {10.1186/s13023-022-02358-x},
	i_language = {en},
	number = {1},
	journal = {Orphanet Journal of Rare Diseases},
	author = {Bauskis, Alicia and Strange, Cecily and Molster, Caron and Fisher, Colleen},
	month = jun,
	year = {2022},
	pages = {233},
}

@article{faye_time_2024,
	title = {Time to diagnosis and determinants of diagnostic delays of people living with a rare disease: results of a {Rare} {Barometer} retrospective patient survey},
	volume = {32},
	issn = {1476-5438},
	i_url = {https://doi.org/10.1038/s41431-024-01604-z},
	doi = {10.1038/s41431-024-01604-z},
	i_language = {en},
	number = {9},
	journal = {European Journal of Human Genetics},
	author = {Faye, Fatoumata and Crocione, Claudia and Anido de Peña, Roberta and Bellagambi, Simona and Escati Peñaloza, Luciana and Hunter, Amy and Jensen, Lene and Oosterwijk, Cor and Schoeters, Eva and de Vicente, Daniel and Faivre, Laurence and Wilbur, Michael and Le Cam, Yann and Dubief, Jessie},
	month = sep,
	year = {2024},
	pages = {1116--1126},
}

@article{michaels-igbokwe_standardized_2021,
	title = {({Un})standardized testing: the diagnostic odyssey of children with rare genetic disorders in {Alberta}, {Canada}},
	volume = {23},
	issn = {1098-3600},
	i_url = {https://doi.org/10.1038/s41436-020-00975-0},
	doi = {10.1038/s41436-020-00975-0},
	i_language = {en},
	number = {2},
	urldate = {2024-05-05},
	journal = {Genetics in Medicine},
	i_publisher = {Elsevier},
	author = {Michaels-Igbokwe, Christine and McInnes, Brenda and MacDonald, Karen V. and Currie, Gillian R. and Omar, Fadya and Shewchuk, Brittany and Bernier, Francois P. and Marshall, Deborah A.},
	month = feb,
	year = {2021},
	pages = {272--279},
}

@article{yang_rdmaster_2024,
	title = {{RDmaster}: {A} novel phenotype-oriented dialogue system supporting differential diagnosis of rare disease},
	volume = {169},
	issn = {0010-4825},
	i_url = {https://www.sciencedirect.com/science/article/pii/S0010482524000088},
	doi = {10.1016/j.compbiomed.2024.107924},
	i_language = {en},
	journal = {Computers in Biology and Medicine},
	author = {Yang, Jian and Shu, Liqi and Han, Mingyu and Pan, Jiarong and Chen, Lihua and Yuan, Tianming and Tan, Linhua and Shu, Qiang and Duan, Huilong and Li, Haomin},
	month = feb,
	year = {2024},
	pages = {107924},
}

@article{topel_ramedis_2006,
	title = {{RAMEDIS}: the rare metabolic diseases database},
	volume = {5},
	issn = {1175-5636},
	i_url = {https://doi.org/10.2165/00822942-200605020-00006},
	doi = {10.2165/00822942-200605020-00006},
	i_language = {en},
	number = {2},
	journal = {Applied Bioinformatics},
	author = {Töpel, Thoralf and Hofestädt, Ralf and Scheible, Dagmar and Trefz, Friedrich},
	month = jun,
	year = {2006},
	pages = {115--118},
}

@article{robinson_interpretable_2020,
	title = {Interpretable {Clinical} {Genomics} with a {Likelihood} {Ratio} {Paradigm}},
	volume = {107},
	issn = {15376605},
	doi = {10.1016/j.ajhg.2020.06.021},
	i_language = {en},
	number = {3},
	journal = {American Journal of Human Genetics},
	i_publisher = {Cell Press},
	author = {Robinson, Peter N. and Ravanmehr, Vida and Jacobsen, Julius O. B. and Danis, Daniel and Zhang, Xingmin Aaron and Carmody, Leigh C. and Gargano, Michael A. and Thaxton, Courtney L. and Karlebach, Guy and Reese, Justin and Holtgrewe, Manuel and Köhler, Sebastian and McMurry, Julie A. and Haendel, Melissa A. and Smedley, Damian},
	month = sep,
	year = {2020},
	pages = {403--417},
}

@article{wakap_estimating_2020,
	title = {Estimating cumulative point prevalence of rare diseases: analysis of the {Orphanet} database},
	volume = {28},
	issn = {14765438},
	doi = {10.1038/s41431-019-0508-0},
	i_language = {en},
	number = {2},
	journal = {European Journal of Human Genetics},
	i_publisher = {Springer Nature},
	author = {Wakap, Stéphanie Nguengang and Lambert, Deborah M. and Olry, Annie and Rodwell, Charlotte and Gueydan, Charlotte and Lanneau, Valérie and Murphy, Daniel and Cam, Yann Le and Rath, Ana},
	month = feb,
	year = {2020},
	pages = {165--173},
}

@article{yang_rdguru_2025,
	title = {{RDguru}: {A} {Conversational} {Intelligent} {Agent} for {Rare} {Diseases}},
	volume = {29},
	doi = {10.1109/JBHI.2024.3464555},
	i_language = {en},
	number = {9},
	journal = {IEEE Journal of Biomedical and Health Informatics},
	author = {Yang, Jian and Shu, Liqi and Duan, Huilong and Li, Haomin},
	year = {2025},
	pages = {6366--6378},
}

@article{chen_rarearena_2026,
	title = {{RareArena}: a comprehensive benchmark dataset unveiling the potential of large language models in rare disease diagnosis},
	volume = {8},
	issn = {2589-7500},
	i_url = {https://doi.org/10.1016/j.landig.2025.100953},
	doi = {10.1016/j.landig.2025.100953},
	i_language = {en},
	number = {2},
	urldate = {2026-05-18},
	journal = {The Lancet Digital Health},
	i_publisher = {Elsevier},
	author = {Chen, Haichao and Zhao, Zhengyun and Zhou, Songchi and Hu, Shikai and Wang, Jinyuan and Jin, Ye and Jin, Xianghong and Tham, Yih Chung and Wang, Xiaofei and Ma, Weizhi and Wu, Honghan and Sheng, Bin and Zhang, Shuyang and Yu, Sheng and Wong, Tien Yin},
	month = feb,
	year = {2026},
}

@article{zhao_agentic_2026,
	title = {An agentic system for rare disease diagnosis with traceable reasoning},
	volume = {651},
	issn = {1476-4687},
	i_url = {https://doi.org/10.1038/s41586-025-10097-9},
	doi = {10.1038/s41586-025-10097-9},
	i_language = {en},
	number = {8106},
	journal = {Nature},
	author = {Zhao, Weike and Wu, Chaoyi and Fan, Yanjie and Qiu, Pengcheng and Zhang, Xiaoman and Sun, Yuze and Zhou, Xiao and Zhang, Shuju and Peng, Yu and Wang, Yanfeng and Sun, Xin and Zhang, Ya and Yu, Yongguo and Sun, Kun and Xie, Weidi},
	month = mar,
	year = {2026},
	pages = {775--784},
}

@article{danis_corpus_2025,
	title = {A corpus of {GA4GH} phenopackets: {Case}-level phenotyping for genomic diagnostics and discovery},
	volume = {6},
	issn = {2666-2477},
	i_url = {https://doi.org/10.1016/j.xhgg.2024.100371},
	doi = {10.1016/j.xhgg.2024.100371},
	i_language = {en},
	number = {1},
	urldate = {2026-05-18},
	journal = {Human Genetics and Genomics Advances},
	i_publisher = {Elsevier},
	author = {Danis, Daniel and Bamshad, Michael J. and Bridges, Yasemin and Caballero-Oteyza, Andrés and Cacheiro, Pilar and Carmody, Leigh C. and Chimirri, Leonardo and Chong, Jessica X. and Coleman, Ben and Dalgleish, Raymond and Freeman, Peter J. and Graefe, Adam S.L. and Groza, Tudor and Hansen, Peter and Jacobsen, Julius O.B. and Klocperk, Adam and Kusters, Maaike and Ladewig, Markus S. and Marcello, Allison J. and Mattina, Teresa and Mungall, Christopher J. and Munoz-Torres, Monica C. and Reese, Justin T. and Rehburg, Filip and Reis, Bárbara C.S. and Schuetz, Catharina and Smedley, Damian and Strauss, Timmy and Sundaramurthi, Jagadish Chandrabose and Thun, Sylvia and Wissink, Kyran and Wagstaff, John F. and Zocche, David and Haendel, Melissa A. and Robinson, Peter N.},
	month = jan,
	year = {2025},
	pages = {100371}
}

@article{philippakis_matchmaker_2015,
	title = {The {Matchmaker} {Exchange}: {A} {Platform} for {Rare} {Disease} {Gene} {Discovery}},
	volume = {36},
	issn = {1059-7794},
	i_url = {https://doi.org/10.1002/humu.22858},
	doi = {10.1002/humu.22858},
	i_language = {en},
	number = {10},
	urldate = {2026-05-18},
	journal = {Human Mutation},
	i_publisher = {John Wiley \& Sons, Ltd},
	author = {Philippakis, Anthony A. and Azzariti, Danielle R. and Beltran, Sergi and Brookes, Anthony J. and Brownstein, Catherine A. and Brudno, Michael and Brunner, Han G. and Buske, Orion J. and Carey, Knox and Doll, Cassie and Dumitriu, Sergiu and Dyke, Stephanie O.M. and den Dunnen, Johan T. and Firth, Helen V. and Gibbs, Richard A. and Girdea, Marta and Gonzalez, Michael and Haendel, Melissa A. and Hamosh, Ada and Holm, Ingrid A. and Huang, Lijia and Hurles, Matthew E. and Hutton, Ben and Krier, Joel B. and Misyura, Andriy and Mungall, Christopher J. and Paschall, Justin and Paten, Benedict and Robinson, Peter N. and Schiettecatte, François and Sobreira, Nara L. and Swaminathan, Ganesh J. and Taschner, Peter E. and Terry, Sharon F. and Washington, Nicole L. and Züchner, Stephan and Boycott, Kym M. and Rehm, Heidi L.},
	month = oct,
	year = {2015},
	pages = {915--921},
}

@article{ronicke_can_2019,
	title = {Can a decision support system accelerate rare disease diagnosis? {Evaluating} the potential impact of {Ada} {DX} in a retrospective study},
	volume = {14},
	issn = {1750-1172},
	i_url = {https://doi.org/10.1186/s13023-019-1040-6},
	doi = {10.1186/s13023-019-1040-6},
	i_language = {en},
	number = {1},
	journal = {Orphanet Journal of Rare Diseases},
	author = {Ronicke, Simon and Hirsch, Martin C. and Türk, Ewelina and Larionov, Katharina and Tientcheu, Daphne and Wagner, Annette D.},
	month = mar,
	year = {2019},
	pages = {69},
}

@article{stark_clinically_2017,
	title = {A clinically driven variant prioritization framework outperforms purely computational approaches for the diagnostic analysis of singleton {WES} data},
	volume = {25},
	issn = {1476-5438},
	i_url = {https://doi.org/10.1038/ejhg.2017.123},
	doi = {10.1038/ejhg.2017.123},
	i_language = {en},
	number = {11},
	journal = {European Journal of Human Genetics},
	author = {Stark, Zornitza and Dashnow, Harriet and Lunke, Sebastian and Tan, Tiong Y and Yeung, Alison and Sadedin, Simon and Thorne, Natalie and Macciocca, Ivan and Gaff, Clara and Oshlack, Alicia and White, Susan M and James, Paul A and {Melbourne Genomics Health Alliance}},
	month = nov,
	year = {2017},
	pages = {1268--1272},
}

@article{wen_phenodp_2025,
	title = {{PhenoDP}: leveraging deep learning for phenotype-based case reporting, disease ranking, and symptom recommendation},
	volume = {17},
	issn = {1756-994X},
	i_url = {https://doi.org/10.1186/s13073-025-01496-8},
	doi = {10.1186/s13073-025-01496-8},
	i_language = {en},
	number = {1},
	journal = {Genome Medicine},
	author = {Wen, Baole and Shi, Sheng and Long, Yi and Dang, Yanan and Tian, Weidong},
	month = jun,
	year = {2025},
	pages = {67},
}

@article{mao_phenotype-based_2025,
	title = {A phenotype-based {AI} pipeline outperforms human experts in differentially diagnosing rare diseases using {EHRs}},
	volume = {8},
	issn = {2398-6352},
	i_url = {https://doi.org/10.1038/s41746-025-01452-1},
	doi = {10.1038/s41746-025-01452-1},
	i_language = {en},
	number = {1},
	journal = {npj Digital Medicine},
	author = {Mao, Xiaohao and Huang, Yu and Jin, Ye and Wang, Lun and Chen, Xuanzhong and Liu, Honghong and Yang, Xinglin and Xu, Haopeng and Luan, Xiaodong and Xiao, Ying and Feng, Siqin and Zhu, Jiahao and Zhang, Xuegong and Jiang, Rui and Zhang, Shuyang and Chen, Ting},
	month = jan,
	year = {2025},
	pages = {68},
}

@article{remy_biolord-2023_2024,
	title = {{BioLORD}-2023: semantic textual representations fusing large language models and clinical knowledge graph insights},
	volume = {31},
	issn = {1527-974X},
	i_url = {https://doi.org/10.1093/jamia/ocae029},
	doi = {10.1093/jamia/ocae029},
	i_language = {en},
	number = {9},
	urldate = {2026-05-18},
	journal = {Journal of the American Medical Informatics Association},
	author = {Remy, François and Demuynck, Kris and Demeester, Thomas},
	month = sep,
	year = {2024},
	pages = {1844--1855},
}

@article{koehn_statistical_2004,
	i_address = {Barcelona, Spain},
	title = {Statistical {Significance} {Tests} for {Machine} {Translation} {Evaluation}},
	i_url = {https://aclanthology.org/W04-3250/},
	i_language = {en},
	journal = {Proceedings of the 2004 {Conference} on {Empirical} {Methods} in {Natural} {Language} {Processing}},
	i_publisher = {Association for Computational Linguistics},
	author = {Koehn, Philipp},
	editor = {Lin, Dekang and Wu, Dekai},
	month = jul,
	year = {2004},
	pages = {388--395},
}

@article{gargano_human_2024,
	title = {The {Human} {Phenotype} {Ontology} in 2024: phenotypes around the world},
	volume = {52},
	issn = {0305-1048},
	i_url = {https://doi.org/10.1093/nar/gkad1005},
	doi = {10.1093/nar/gkad1005},
	i_language = {en},
	number = {D1},
	urldate = {2026-05-19},
	journal = {Nucleic Acids Research},
	author = {Gargano, Michael A and Matentzoglu, Nicolas and Coleman, Ben and Addo-Lartey, Eunice B and Anagnostopoulos, Anna V and Anderton, Joel and Avillach, Paul and Bagley, Anita M and Bakštein, Eduard and Balhoff, James P and Baynam, Gareth and Bello, Susan M and Berk, Michael and Bertram, Holli and Bishop, Somer and Blau, Hannah and Bodenstein, David F and Botas, Pablo and Boztug, Kaan and Čady, Jolana and Callahan, Tiffany J and Cameron, Rhiannon and Carbon, Seth J and Castellanos, Francisco and Caufield, J Harry and Chan, Lauren E and Chute, Christopher G and Cruz-Rojo, Jaime and Dahan-Oliel, Noémi and Davids, Jon R and de Dieuleveult, Maud and de Souza, Vinicius and de Vries, Bert B A and de Vries, Esther and DePaulo, J Raymond and Derfalvi, Beata and Dhombres, Ferdinand and Diaz-Byrd, Claudia and Dingemans, Alexander J M and Donadille, Bruno and Duyzend, Michael and Elfeky, Reem and Essaid, Shahim and Fabrizzi, Carolina and Fico, Giovanna and Firth, Helen V and Freudenberg-Hua, Yun and Fullerton, Janice M and Gabriel, Davera L and Gilmour, Kimberly and Giordano, Jessica and Goes, Fernando S and Moses, Rachel Gore and Green, Ian and Griese, Matthias and Groza, Tudor and Gu, Weihong and Guthrie, Julia and Gyori, Benjamin and Hamosh, Ada and Hanauer, Marc and Hanušová, Kateřina and He, Yongqun (Oliver) and Hegde, Harshad and Helbig, Ingo and Holasová, Kateřina and Hoyt, Charles Tapley and Huang, Shangzhi and Hurwitz, Eric and Jacobsen, Julius O B and Jiang, Xiaofeng and Joseph, Lisa and Keramatian, Kamyar and King, Bryan and Knoflach, Katrin and Koolen, David A and Kraus, Megan L and Kroll, Carlo and Kusters, Maaike and Ladewig, Markus S and Lagorce, David and Lai, Meng-Chuan and Lapunzina, Pablo and Laraway, Bryan and Lewis-Smith, David and Li, Xiarong and Lucano, Caterina and Majd, Marzieh and Marazita, Mary L and Martinez-Glez, Victor and McHenry, Toby H and McInnis, Melvin G and McMurry, Julie A and Mihulová, Michaela and Millett, Caitlin E and Mitchell, Philip B and Moslerová, Veronika and Narutomi, Kenji and Nematollahi, Shahrzad and Nevado, Julian and Nierenberg, Andrew A and Čajbiková, Nikola Novák and Nurnberger, Jr., John I and Ogishima, Soichi and Olson, Daniel and Ortiz, Abigail and Pachajoa, Harry and Perez de Nanclares, Guiomar and Peters, Amy and Putman, Tim and Rapp, Christina K and Rath, Ana and Reese, Justin and Rekerle, Lauren and Roberts, Angharad M and Roy, Suzy and Sanders, Stephan J and Schuetz, Catharina and Schulte, Eva C and Schulze, Thomas G and Schwarz, Martin and Scott, Katie and Seelow, Dominik and Seitz, Berthold and Shen, Yiping and Similuk, Morgan N and Simon, Eric S and Singh, Balwinder and Smedley, Damian and Smith, Cynthia L and Smolinsky, Jake T and Sperry, Sarah and Stafford, Elizabeth and Stefancsik, Ray and Steinhaus, Robin and Strawbridge, Rebecca and Sundaramurthi, Jagadish Chandrabose and Talapova, Polina and Tenorio Castano, Jair A and Tesner, Pavel and Thomas, Rhys H and Thurm, Audrey and Turnovec, Marek and van Gijn, Marielle E and Vasilevsky, Nicole A and Vlčková, Markéta and Walden, Anita and Wang, Kai and Wapner, Ron and Ware, James S and Wiafe, Addo A and Wiafe, Samuel A and Wiggins, Lisa D and Williams, Andrew E and Wu, Chen and Wyrwoll, Margot J and Xiong, Hui and Yalin, Nefize and Yamamoto, Yasunori and Yatham, Lakshmi N and Yocum, Anastasia K and Young, Allan H and Yüksel, Zafer and Zandi, Peter P and Zankl, Andreas and Zarante, Ignacio and Zvolský, Miroslav and Toro, Sabrina and Carmody, Leigh C and Harris, Nomi L and Munoz-Torres, Monica C and Danis, Daniel and Mungall, Christopher J and Köhler, Sebastian and Haendel, Melissa A and Robinson, Peter N},
	month = jan,
	year = {2024},
	pages = {D1333--D1346},
}

@article{cheng_survey_2025,
	title = {A {Survey} on {Data} {Contamination} for {Large} {Language} {Models}},
	i_url = {https://arxiv.org/abs/2502.14425},
	journal = {arXiv preprint arXiv:2502.14425},
	author = {Cheng, Yuxing and Chang, Yi and Wu, Yuan},
	year = {2025},
	i_note = {\_eprint: 2502.14425},
}

@article{ghassemi_false_2021,
	title = {The false hope of current approaches to explainable artificial intelligence in health care},
	volume = {3},
	issn = {2589-7500},
	i_url = {https://doi.org/10.1016/S2589-7500(21)00208-9},
	doi = {10.1016/S2589-7500(21)00208-9},
	i_language = {en},
	number = {11},
	urldate = {2026-05-18},
	journal = {The Lancet Digital Health},
	i_publisher = {Elsevier},
	author = {Ghassemi, Marzyeh and Oakden-Rayner, Luke and Beam, Andrew L},
	month = nov,
	year = {2021},
	pages = {e745--e750},
}

@article{the_lancet_global_health_landscape_2024,
	title = {The landscape for rare diseases in 2024},
	volume = {12},
	issn = {2214-109X},
	i_url = {https://doi.org/10.1016/S2214-109X(24)00056-1},
	doi = {10.1016/S2214-109X(24)00056-1},
	number = {3},
	urldate = {2026-05-22},
	journal = {The Lancet Global Health},
	i_publisher = {Elsevier},
	author = {{The Lancet Global Health}},
	month = mar,
	year = {2024},
	pages = {e341},
}

@article{chen_rarebench_2024,
	title = {{RareBench}: {Can} {LLMs} {Serve} as {Rare} {Diseases} {Specialists}?},
	i_url = {https://arxiv.org/abs/2402.06341},
	doi = {10.48550/arXiv.2402.06341},
	i_language = {en},
	journal = {ACM SIGKDD Conference on Knowledge Discovery and Data Mining},
	author = {Chen, Xuanzhong and Mao, Xiaohao and Guo, Qihan and Wang, Lun and Zhang, Shuyang and Chen, Ting},
	year = {2024},
	i_note = {\_eprint: 2402.06341},
}

@article{vaswani_attention_2017,
	title = {Attention is all you need},
	volume = {30},
	i_url = {https://proceedings.neurips.cc/paper_files/paper/2017/file/3f5ee243547dee91fbd053c1c4a845aa-Paper.pdf},
	i_language = {en},
	journal = {Proceedings of the 31st International Conference on Neural Information Processing Systems},
	author = {Vaswani, Ashish and Shazeer, Noam and Parmar, Niki and Uszkoreit, Jakob and Jones, Llion and Gomez, Aidan N. and Kaiser, Ł ukasz and Polosukhin, Illia},
	editor = {Guyon, I. and Luxburg, U. Von and Bengio, S. and Wallach, H. and Fergus, R. and Vishwanathan, S. and Garnett, R.},
	year = {2017},
	pages = {5998--6008},
}

@article{yang_specialized_2025,
	title = {A {Specialized} {Large} {Language} {Model} for {Clinical} {Reasoning} and {Diagnosis} in {Rare} {Diseases}},
	i_url = {https://arxiv.org/abs/2511.14638},
	i_language = {en},
	journal = {arXiv preprint arXiv:2511.14638},
	author = {Yang, Tao and Huang, Dandan and Lin, Yunting and Wu, Pengfei and Wu, Zhikun and Ma, Gangyuan and Lu, Yulan and Dong, Xinran and Li, Dingpeng and Ge, Junshuang and Zhang, Zhiyan and Huang, Xuanzhao and Nong, Wenyan and Zhou, Yao and Tang, Hui and Yang, Hongxi and Zhang, Shijie and Li, Juan and Cao, Xiaojun and Yang, Lin and Gao, Xia and Xu, Kaishou and Gu, Xiaoqiong and Zhang, Wen and Xia, Huimin and Liu, Li and Zhou, Wenhao and Li, Mulin Jun},
	year = {2025},
	i_note = {\_eprint: 2511.14638},
}

@article{chen_xgboost_2016,
	i_address = {New York, NY, USA},
	i_series = {{KDD} '16},
	title = {{XGBoost}: {A} {Scalable} {Tree} {Boosting} {System}},
	i_isbn = {978-1-4503-4232-2},
	i_url = {https://doi.org/10.1145/2939672.2939785},
	doi = {10.1145/2939672.2939785},
	i_language = {en},
	journal = {Proceedings of the 22nd {ACM} {SIGKDD} {International} {Conference} on {Knowledge} {Discovery} and {Data} {Mining}},
	i_publisher = {Association for Computing Machinery},
	author = {Chen, Tianqi and Guestrin, Carlos},
	editor = {Lin, Dekang and Wu, Dekai},
	year = {2016},
	pages = {785--794},
}

@article{goldenberg_is_2026,
	title = {Is {AI} actually improving healthcare?},
	volume = {32},
	issn = {1546-170X},
	i_url = {https://doi.org/10.1038/s41591-026-04329-2},
	doi = {10.1038/s41591-026-04329-2},
	i_language = {en},
	number = {4},
	journal = {Nature Medicine},
	author = {Goldenberg, Anna and Wiens, Jenna},
	month = apr,
	year = {2026},
	pages = {1182--1183},
}

@article{pedregosa_scikit-learn_2011,
	title = {Scikit-learn: {Machine} {Learning} in {Python}},
	volume = {12},
	issn = {1532-4435},
	journal = {The Journal of Machine Learning Research},
	author = {Pedregosa, Fabian and Varoquaux, Gaël and Gramfort, Alexandre and Michel, Vincent and Thirion, Bertrand and Grisel, Olivier and Blondel, Mathieu and Prettenhofer, Peter and Weiss, Ron and Dubourg, Vincent and Vanderplas, Jake and Passos, Alexandre and Cournapeau, David and Brucher, Matthieu and Perrot, Matthieu and Duchesnay, Édouard},
	month = nov,
	year = {2011},
	pages = {2825--2830},
}

@article{rumelhart_learning_1986,
	title = {Learning representations by back-propagating errors},
	volume = {323},
	issn = {1476-4687},
	i_url = {https://doi.org/10.1038/323533a0},
	doi = {10.1038/323533a0},
	i_language = {en},
	number = {6088},
	journal = {Nature},
	author = {Rumelhart, David E. and Hinton, Geoffrey E. and Williams, Ronald J.},
	month = oct,
	year = {1986},
	pages = {533--536},
}

@article{omar_how_2026,
	title = {How to meaningfully evaluate {AI} in clinical medicine},
	volume = {32},
	issn = {1546-170X},
	i_url = {https://doi.org/10.1038/s41591-026-04350-5},
	doi = {10.1038/s41591-026-04350-5},
	i_language = {en},
	number = {6},
	journal = {Nature Medicine},
	author = {Omar, Mahmud and Agbareia, Reem and Gorenshtein, Alon and Ramaswamy, Ashwin and Sakhuja, Ankit and Barash, Yiftach and Ting, Daniel S. W. and Klang, Eyal and Nadkarni, Girish N.},
	month = jun,
	year = {2026},
	pages = {1948--1949},
}

@article{zhao_large-scale_2023,
	title = {A large-scale dataset of patient summaries for retrieval-based clinical decision support systems},
	volume = {10},
	issn = {2052-4463},
	i_url = {https://doi.org/10.1038/s41597-023-02814-8},
	doi = {10.1038/s41597-023-02814-8},
	number = {1},
	journal = {Scientific Data},
	author = {Zhao, Zhengyun and Jin, Qiao and Chen, Fangyuan and Peng, Tuorui and Yu, Sheng},
	month = dec,
	year = {2023},
	pages = {909},
}

@article{manda_large_2025,
	title = {Large {Language} {Models} in {Bio}-{Ontology} {Research}: {A} {Review}},
	volume = {12},
	issn = {2306-5354},
	doi = {10.3390/bioengineering12111260},
	number = {11},
	journal = {Bioengineering},
	author = {Manda, Prashanti},
	year = {2025},
	pages = {1260},
}

@article{yang_robust_2023,
	title = {A robust phenotype-driven likelihood ratio analysis approach assisting interpretable clinical diagnosis of rare diseases},
	volume = {142},
	issn = {1532-0464},
	i_url = {https://www.sciencedirect.com/science/article/pii/S153204642300093X},
	doi = {10.1016/j.jbi.2023.104372},
	i_language = {en},
	journal = {Journal of Biomedical Informatics},
	author = {Yang, Jian and Shu, Liqi and Duan, Huilong and Li, Haomin},
	year = {2023},
	pages = {104372},
}

@article{paszke_pytorch_2019,
	i_address = {Red Hook, NY, USA},
	title = {{PyTorch}: an imperative style, high-performance deep learning library},
	journal = {Proceedings of the 33rd {International} {Conference} on {Neural} {Information} {Processing} {Systems}},
	i_publisher = {Curran Associates Inc.},
	author = {Paszke, Adam and Gross, Sam and Massa, Francisco and Lerer, Adam and Bradbury, James and Chanan, Gregory and Killeen, Trevor and Lin, Zeming and Gimelshein, Natalia and Antiga, Luca and Desmaison, Alban and Köpf, Andreas and Yang, Edward and DeVito, Zach and Raison, Martin and Tejani, Alykhan and Chilamkurthy, Sasank and Steiner, Benoit and Fang, Lu and Bai, Junjie and Chintala, Soumith},
	year = {2019},
}

@article{cho_building_2026,
	title = {Building {Safe} and {Deployable} {Clinical} {Natural} {Language} {Processing} under {Temporal} {Leakage} {Constraints}},
	i_url = {https://arxiv.org/abs/2602.15852},
	journal = {arXiv preprint arXiv:2602.15852},
	author = {Cho, Ha Na and Sutari, Sairam and Lopez, Alexander and Bow, Hansen and Zheng, Kai},
	year = {2026},
	i_note = {\_eprint: 2602.15852},
}

@article{kapoor_leakage_2023,
	title = {Leakage and the reproducibility crisis in machine-learning-based science},
	volume = {4},
	i_url = {https://www.sciencedirect.com/science/article/pii/S2666389923001599},
	doi = {10.1016/j.patter.2023.100804},
	number = {9},
	journal = {Patterns},
	author = {Kapoor, Sayash and Narayanan, Arvind},
	month = sep,
	year = {2023},
	pages = {100804},
}

@article{moher_consort_2010,
	title = {{CONSORT} 2010 {Explanation} and {Elaboration}: updated guidelines for reporting parallel group randomised trials},
	volume = {340},
	i_url = {https://www.bmj.com/content/340/bmj.c869.abstract},
	doi = {10.1136/bmj.c869},
	journal = {BMJ},
	author = {Moher, David and Hopewell, Sally and Schulz, Kenneth F and Montori, Victor and Gøtzsche, Peter C and Devereaux, P J and Elbourne, Diana and Egger, Matthias and Altman, Douglas G},
	month = mar,
	year = {2010},
	pages = {c869},
}

@article{liu_reporting_2020,
	title = {Reporting guidelines for clinical trial reports for interventions involving artificial intelligence: the {CONSORT}-{AI} extension},
	volume = {26},
	i_url = {https://doi.org/10.1038/s41591-020-1034-x},
	doi = {10.1038/s41591-020-1034-x},
	number = {9},
	journal = {Nature Medicine},
	author = {Liu, Xiaoxuan and Cruz Rivera, Samantha and Moher, David and Calvert, Melanie J. and Denniston, Alastair K. and Chan, An-Wen and Darzi, Ara and Holmes, Christopher and Yau, Christopher and Ashrafian, Hutan and Deeks, Jonathan J. and Ferrante di Ruffano, Lavinia and Faes, Livia and Keane, Pearse A. and Vollmer, Sebastian J. and Lee, Aaron Y. and Jonas, Adrian and Esteva, Andre and Beam, Andrew L. and Chan, An-Wen and Panico, Maria Beatrice and Lee, Cecilia S. and Haug, Charlotte and Kelly, Christopher J. and Yau, Christopher and Mulrow, Cynthia and Espinoza, Cyrus and Fletcher, John and Paltoo, Dina and Manna, Elaine and Price, Gary and Collins, Gary S. and Harvey, Hugh and Matcham, James and Monteiro, Joao and ElZarrad, M. Khair and Ferrante di Ruffano, Lavinia and Oakden-Rayner, Luke and McCradden, Melissa and Keane, Pearse A. and Savage, Richard and Golub, Robert and Sarkar, Rupa and Rowley, Samuel and {The SPIRIT-AI and CONSORT-AI Working Group} and {SPIRIT-AI and CONSORT-AI Steering Group} and {SPIRIT-AI and CONSORT-AI Consensus Group}},
	month = sep,
	year = {2020},
	pages = {1364--1374},
}

@article{yang_fusionrare_2026,
	author    = {Yang, Jian},
	title     = {FusionRare's challenge cases},
	year      = {2026},
	journal   = {Zenodo},
	doi       = {10.5281/zenodo.20697776},
	url       = {https://doi.org/10.5281/zenodo.20697776},
	note      = {\url{https://doi.org/10.5281/zenodo.20697776}},
}

@article{zheng_end--end_2025,
	title = {End-to-{End} {Agentic} {RAG} {System} {Training} for {Traceable} {Diagnostic} {Reasoning}},
	i_url = {https://arxiv.org/abs/2508.15746},
	author = {Zheng, Qiaoyu and Sun, Yuze and Wu, Chaoyi and Zhao, Weike and Qiu, Pengcheng and Yu, Yongguo and Sun, Kun and Zhang, Jian and Wang, Yanfeng and Zhang, Ya and Xie, Weidi},
	journal = {arXiv preprint arXiv:2508.15746},
	year = {2025},
}

@article{rose_meddxagent_2025,
	title = {{MEDDxAgent}: {A} {Unified} {Modular} {Agent} {Framework} for {Explainable} {Automatic} {Differential} {Diagnosis}},
	i_url = {https://arxiv.org/abs/2502.19175},
	author = {Rose, Daniel and Hung, Chia-Chien and Lepri, Marco and Alqassem, Israa and Gashteovski, Kiril and Lawrence, Carolin},
	year = {2025},
	journal = {arXiv preprint arXiv:2502.19175},
}

@article{schumacher_rare_2025,
	title = {Rare {Disease} {Differential} {Diagnosis} with {Large} {Language} {Models} at {Scale}: {From} {Abdominal} {Actinomycosis} to {Wilson}'s {Disease}},
	i_url = {https://arxiv.org/abs/2502.15069},
	author = {Schumacher, Elliot and Naik, Dhruv and Kannan, Anitha},
	year = {2025},
	journal = {arXiv preprint arXiv:2502.15069},
}

@article{wang_multimodal_2026,
	title = {Multimodal integrated knowledge transfer to large language models through preference optimization with biomedical applications},
	volume = {7},
	issn = {2666-3899},
	i_url = {https://www.sciencedirect.com/science/article/pii/S266638992600053X},
	doi = {10.1016/j.patter.2026.101544},
	number = {8},
	journal = {Patterns},
	author = {Wang, Zhanliang and Wu, Da and Nguyen, Quan M. and Xu, Zhuoran and Wang, Kai},
	month = aug,
	year = {2026},
	pages = {101544},
}

@article{chen_enhancing_2025,
	title = {Enhancing diagnostic capability with multi-agents conversational large language models},
	volume = {8},
	issn = {2398-6352},
	i_url = {https://doi.org/10.1038/s41746-025-01550-0},
	doi = {10.1038/s41746-025-01550-0},
	number = {1},
	journal = {npj Digital Medicine},
	author = {Chen, Xi and Yi, Huahui and You, Mingke and Liu, WeiZhi and Wang, Li and Li, Hairui and Zhang, Xue and Guo, Yingman and Fan, Lei and Chen, Gang and Lao, Qicheng and Fu, Weili and Li, Kang and Li, Jian},
	month = mar,
	year = {2025},
	pages = {159},
}

@article{chen_rareagents_2026,
	title = {{RareAgents}: autonomous multi-disciplinary team for rare disease diagnosis and treatment},
	doi = {10.1609/aaai.v40i1.36969},
	journal = {Proceedings of the {Fortieth} {AAAI} {Conference} on {Artificial} {Intelligence} and {Thirty}-{Eighth} {Conference} on {Innovative} {Applications} of {Artificial} {Intelligence} and {Sixteenth} {Symposium} on {Educational} {Advances} in {Artificial} {Intelligence}},
	author = {Chen, Xuanzhong and Jin, Ye and Mao, Xiaohao and Wang, Lun and Zhang, Shuyang and Chen, Ting},
	year = {2026},
}

@article{nguyen_diagnostic_2026,
	title = {Diagnostic accuracy of large language models for rare diseases: a systematic review and meta-analysis},
	issn = {2398-6352},
	i_url = {https://doi.org/10.1038/s41746-026-03109-z},
	doi = {10.1038/s41746-026-03109-z},
	journal = {npj Digital Medicine},
	author = {Nguyen, Minh-Ha and Yang, Chih-Ting and Cassini, Thomas A. and Ma, Fan and Hamid, Rizwan and Bastarache, Lisa and Peterson, Josh F. and Xu, Hua and Li, Lingyao and Ma, Siyuan and Shyr, Cathy},
	month = aug,
	year = {2026},
}

@article{sounderajah_stard-ai_2025,
	title = {The {STARD}-{AI} reporting guideline for diagnostic accuracy studies using artificial intelligence},
	volume = {31},
	issn = {1546-170X},
	i_url = {https://doi.org/10.1038/s41591-025-03953-8},
	doi = {10.1038/s41591-025-03953-8},
	number = {10},
	journal = {Nature Medicine},
	author = {Sounderajah, Viknesh and Guni, Ahmad and Liu, Xiaoxuan and Collins, Gary S. and Karthikesalingam, Alan and Markar, Sheraz R. and Golub, Robert M. and Denniston, Alastair K. and Shetty, Shravya and Moher, David and Bossuyt, Patrick M. and Darzi, Ara and Ashrafian, Hutan and Acharya, Amish and Mateen, Bilal A. and Kelly, Christopher and Ting, Daniel and Treanor, Darren and King, Dominic and Greaves, Felix and Harvey, Hugh and De Fauw, Jeffrey and Cohen, Jérémie F. and Godwin, Jonathan and Pearson-Stuttard, Jonathan and Moons, Karel and Harling, Leanne and Maier-Hein, Lena and Hooft, Lotty and McInnes, Matthew DF and McCradden, Melissa and Rifai, Nader and Tomasev, Nenad and Normahani, Pasha and Whiting, Penny and Aggarwal, Ravi and Vollmer, Sebastian and Markar, Sheraz R. and Panch, Trishan and {Ben Glocker} and Taylor, David and Moher, David and Samuel, Diana and Ordish, Johan and Singh, Karandeep and Celi, Leo and Bossuyt, Patrick and Rose, Sherri and Saria, Suchi and {STARD-AI Steering Committee} and {STARD-AI Consensus Group}},
	month = oct,
	year = {2025},
	pages = {3283--3289},
}

@article{haendel_how_2020,
	address = {England},
	title = {How many rare diseases are there?},
	volume = {19},
	issn = {1474-1784 1474-1776},
	doi = {10.1038/d41573-019-00180-y},
	language = {eng},
	number = {2},
	journal = {Nature reviews. Drug discovery},
	author = {Haendel, Melissa and Vasilevsky, Nicole and Unni, Deepak and Bologa, Cristian and Harris, Nomi and Rehm, Heidi and Hamosh, Ada and Baynam, Gareth and Groza, Tudor and McMurry, Julie and Dawkins, Hugh and Rath, Ana and Thaxton, Courtney and Bocci, Giovanni and Joachimiak, Marcin P. and Köhler, Sebastian and Robinson, Peter N. and Mungall, Chris and Oprea, Tudor I.},
	month = feb,
	year = {2020},
	pages = {77--78},
}

@article{fujiwara_pubcasefinder_2018,
	title = {{PubCaseFinder}: {A} {Case}-{Report}-{Based}, {Phenotype}-{Driven} {Differential}-{Diagnosis} {System} for {Rare} {Diseases}},
	volume = {103},
	issn = {0002-9297},
	i_url = {https://doi.org/10.1016/j.ajhg.2018.08.003},
	doi = {10.1016/j.ajhg.2018.08.003},
	number = {3},
	urldate = {2026-08-26},
	journal = {The American Journal of Human Genetics},
	publisher = {Elsevier},
	author = {Fujiwara, Toyofumi and Yamamoto, Yasunori and Kim, Jin-Dong and Buske, Orion and Takagi, Toshihisa},
	month = sep,
	year = {2018},
	pages = {389--399},
}


\end{document}


\title[FusionRare]{\textbf{Unified Multiparadigm Fusion Intelligence for Rare Disease Diagnosis}}

\author{
    \fnm{Jian} \sur{Yang}\textsuperscript{1,\dag,*}, 
    \fnm{Peilin} \sur{Liu}\textsuperscript{2,\dag}, 
    \fnm{Songda} \sur{Li}\textsuperscript{2}, 
    \fnm{Yu} \sur{Cao}\textsuperscript{1},
    \fnm{Yuqing} \sur{Feng}\textsuperscript{1},
    \fnm{Lan} \sur{Yu}\textsuperscript{1},
    \fnm{Jiarong} \sur{Pan}\textsuperscript{1},
    \fnm{Jing} \sur{Ye}\textsuperscript{1},
    \fnm{Xinwen} \sur{Huang}\textsuperscript{1},
    \fnm{Chaochun} \sur{Zou}\textsuperscript{1},
    \fnm{Xudong} \sur{Lu}\textsuperscript{2},
    \fnm{Huilong} \sur{Duan}\textsuperscript{2},
    \fnm{and Haomin} \sur{Li}\textsuperscript{1,2,*}
}

\affil[1]{\orgdiv{The Children's Hospital}, \orgname{Zhejiang University School of Medicine}, \orgaddress{\city{Hangzhou}, \postcode{310052}, \country{China}}}

\affil[2]{\orgdiv{The College of Biomedical Engineering and Instrument Science}, \orgname{Zhejiang University}, \orgaddress{\city{Hangzhou}, \postcode{310027}, \country{China}}}

\affil[]{\textsuperscript{\dag}These authors contributed equally to this work}

\affil[]{\textsuperscript{*}Corresponding Author: hmli@zju.edu.cn (H. L.); yangjian\_bme@zju.edu.cn (J. Y.)}

\maketitle

\newpage
\begingroup
\renewcommand{\numberline}[1]{Appendix #1\quad} 
\tableofcontents
\endgroup
\newpage

\begin{appendices}
\input{appendix/A.tex}
\clearpage
\input{appendix/B.tex}
\clearpage
\input{appendix/C.tex}
\clearpage
\input{appendix/D.tex}
\clearpage
\input{appendix/E.tex}
\end{appendices}

\newpage
\renewcommand{\refname}{References for Appendix}
\bibliography{fusionrare} 


\title[]{\textbf{Auditing evaluation artifacts and data leakage in large language model systems for rare disease diagnosis}}

\author{
    \fnm{Jian} \sur{Yang}\textsuperscript{1,*}, 
    \fnm{Peilin} \sur{Liu}\textsuperscript{2}, 
    \fnm{Xiaohan} \sur{Li}\textsuperscript{2},
    \fnm{Songda} \sur{Li}\textsuperscript{2}, 
    \fnm{Yu} \sur{Cao}\textsuperscript{1},
    \fnm{Yuqing} \sur{Feng}\textsuperscript{1},
    \fnm{and Haomin} \sur{Li}\textsuperscript{1,2,*}
}

\affil[1]{\small \orgdiv{The Children's Hospital}, \orgname{Zhejiang University School of Medicine}, \orgaddress{\city{Hangzhou}, \postcode{310052}, \country{China}}}

\affil[2]{\small \orgdiv{The College of Biomedical Engineering and Instrument Science}, \orgname{Zhejiang University}, \orgaddress{\city{Hangzhou}, \postcode{310027}, \country{China}}}

\affil[]{\small \textsuperscript{*}Corresponding Author: hmli@zju.edu.cn (H. L.); yangjian\_bme@zju.edu.cn (J. Y.)}

\maketitle

\clearpage
\section*{Supplementary Information}

\subsection*{Supplementary Text}
\subsubsection*{Prompt settings}

\begin{promptbox}{Prompt 1: Minimalist RAG LLM evaluation with reference cases}
You are a rare disease clinical expert. Please give a ranked list with 10 differential diagnosis names (rare disease names in English only) based on provided target patient information and its reference cases:

\textbf{Target Patient Phenotypes}: [List of HPO Terms]

\textbf{Reference Cases}: [Reference Cases Block, including matched historical cohorts]

\textbf{Note}: Use reference cases for clinical reasoning inspiration only, not for diagnosis copying.

(Give a numerical list of disease names in plain text, without disease descriptions and without other special formatting such as bold)
\end{promptbox}

\begin{promptbox}{Prompt 2: Zero-shot LLM evaluation using clinical narratives}
You are a rare disease clinical expert. Please give a ranked list with 10 differential diagnosis names (rare disease names in English only) based on provided patient information below (patient record may be in Chinese):

\textbf{Chief Complaint}: [Details]

\textbf{Physical Examination}: [Details] (optional)

\textbf{Ancillary Examinations}: [Details] (optional)

\textbf{Clinical Treatment}: [Details] (optional)

\textbf{Supplementary HPOs extracted from records}: [List of HPO Terms] (optional)

(Give a numerical list of disease names in plain text, without disease descriptions and without other special formatting such as bold)
\end{promptbox}

\clearpage
\subsection*{Supplementary Figures}

\setcounter{figure}{0}

\captionsetup[figure]{name=Supplementary Fig.}
\renewcommand{\thefigure}{S\arabic{figure}}

\begin{figure}[ht!]
    \centering
    \includegraphics[width=1.0\textwidth]{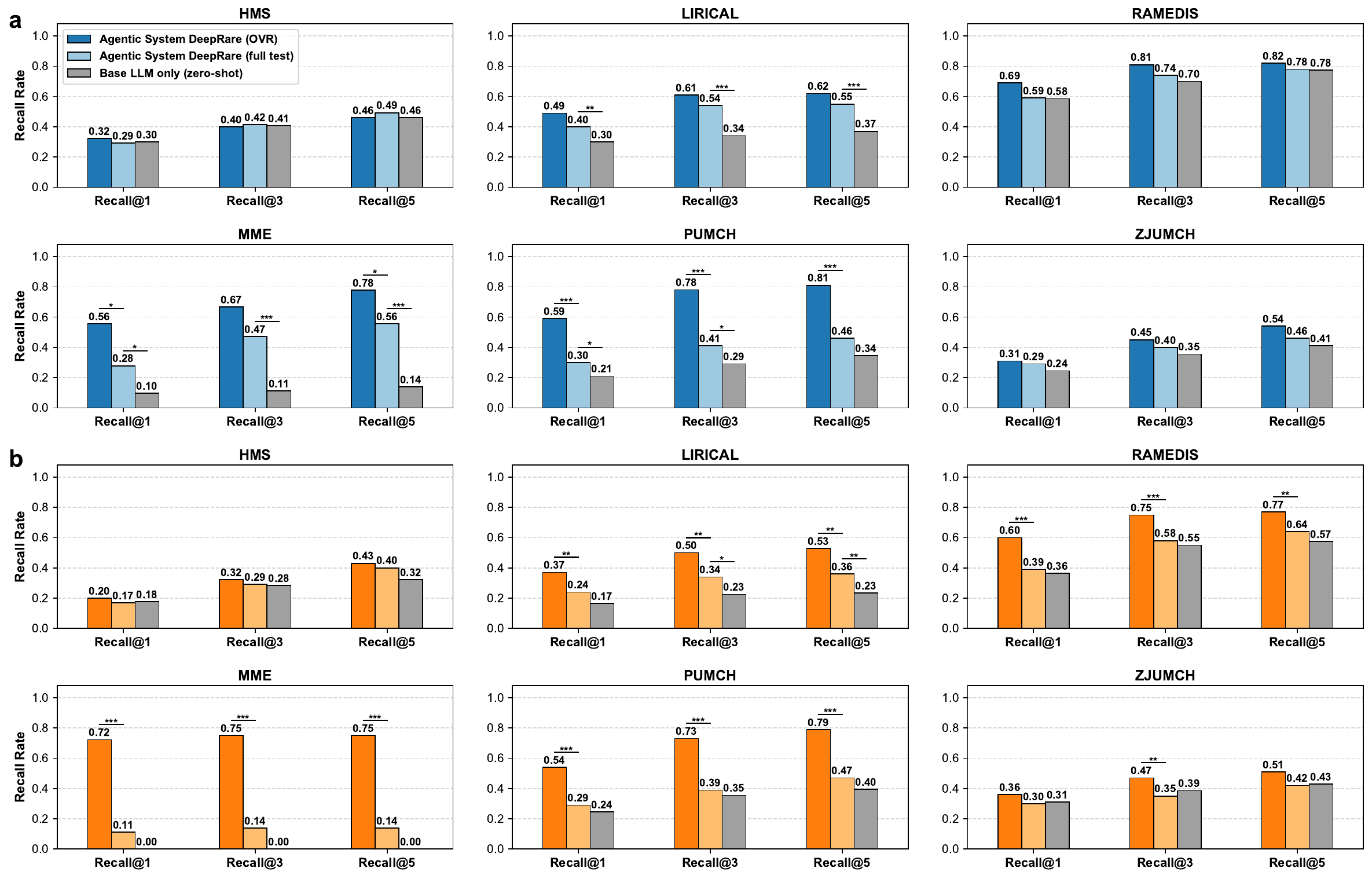} 
    \caption{\textbf{Foundation LLM ablation for DeepRare.}
    Diagnostic recall of DeepRare across six evaluation cohorts when redeployed with (a) Gemini-2.5-Flash and (b) GPT-4o-mini, compared under one-vs-rest (OVR), full-test, and standalone zero-shot settings.}
    \label{fig:deeprare_two_llms}
\end{figure}

\clearpage
\subsection*{Supplementary Tables}
\setcounter{table}{0}
\captionsetup[table]{name=Supplementary Table, position=top}
\renewcommand{\thetable}{S\arabic{table}}

\begingroup
\small
\begin{xltabular}{\textwidth}{
    >{\raggedright\arraybackslash}p{2.2cm} 
    >{\raggedright\arraybackslash}X 
    >{\raggedright\arraybackslash}X 
    >{\raggedright\arraybackslash}X 
    >{\raggedright\arraybackslash}X}

\caption{Detailed evidence and methodological audit across rare disease diagnostic AI systems.} \label{tab:supp_audit_deeprare} \\
\toprule
\textbf{System (year)} & \textbf{Candidate-space Contamination} & \textbf{Spatial Overfitting} & \textbf{Post-hoc Cue Leakage} & \textbf{Temporal Knowledge Bias} \\
\midrule
\endfirsthead

\multicolumn{5}{c}{\small\itshape Cont. Supplementary Table \ref{tab:supp_audit_deeprare}} \\[0.5ex]
\toprule
\textbf{System (year)} & \textbf{Candidate-space Contamination} & \textbf{Spatial Overfitting} & \textbf{Post-hoc Cue Leakage} & \textbf{Temporal Knowledge Bias} \\
\midrule
\endhead
\bottomrule
\endlastfoot

\makecell[l]{\textbf{DeepRare} \\ \textbf{(2026)}} & 
\textbf{[+] High Risk:} The reference case bank ($n=67,795$) retained same-cohort records, applying only single-case exclusion (One-vs-Rest) without dataset-level isolation or PMID-level deduplication for literature cohorts (MME, LIRICAL, RAMEDIS) [p.~14, \textit{Case collection}]. Dynamic web retrieval (PubMed/Google/Bing) further allows direct retrieval of original published case reports [p.~12, \textit{Knowledge searcher}]. & 
\textbf{[--] Low Risk:} DeepRare operates as a training-free agentic framework utilizing frozen foundation LLMs without supervised fine-tuning or parameter updates on specific institutional EHRs, avoiding parameter-level memorization of local center distributions [p.~2, \textit{System overview}; p.~11, \textit{Methods}]. & 
\textbf{[--] Low Risk:} Unstructured narratives are converted to isolated HPO entities via the \textit{Phenotype Extractor}; diagnostic reasoning is executed strictly over structured HPO terms and/or VCF files, structurally filtering out retrospective treatment courses and discharge outcomes [p.~11, \textit{Eqs.~2--6}; p.~12, \textit{Phenotype extractor}]. & 
\textbf{[+] High Risk:} Integrates real-time online literature retrieval and contemporary toolchains (e.g., Exomiser v14.1.0, Nov 2024) while evaluating against offline baselines (e.g., PhenoBrain) fixed to historical KB snapshots, without chronologically synchronizing reference KB versions across baselines [p.~2, \textit{Evaluation settings}; p.~12; p.~15].\\

\makecell[l]{\textbf{RareAgents} \\ \textbf{(2026)}} & 
\textbf{[+] High Risk:} The \textit{Dynamic Long-term Memory} retrieves Top-$5$ similar reference cases directly from the RareBench database ($n=1,197$) to inject into the prompt context without source-level isolation or PMID deduplication [p.~4, \textit{Dynamic Long-term Memory}; p.~20, \textit{Case 1}]. Ablation shows diagnostic Hit@1 plummets by 22.4\% (55.89\% to 43.36\%) upon removing memory retrieval, confirming heavy reliance on candidate-space leakage [p.~6; p.~7, \textit{Table~6}]. & 
\textbf{[--] Low Risk:} Operates as a training-free agent framework using frozen Llama-3.1-8B/70B-Instruct backbones without parameter fine-tuning on institutional EHRs, avoiding parameter-level memorization of local center distributions [p.~2, \textit{Contributions}; p.~10, \textit{Limitations}]. & 
\textbf{[--] Low Risk:} Diagnostic input is strictly restricted to patient symptom/phenotype lists, structurally excluding retrospective unstructured text, therapeutic interventions, invasive procedures, and discharge summaries from diagnostic reasoning [p.~3, \textit{Problem Formulation}; p.~18, \textit{Case 1}]. & 
\textbf{[+] High Risk:} Simultaneously integrates multiple heterogeneous diagnostic tools (Phenomizer, LIRICAL, PhenoBrain) via function calling without clearly specifying the chronological synchronization of reference KB versions across integrated tools and compared baselines [p.~5, \textit{Medical Tools Utilization}; p.~11, \textit{Medical tools}].\\

\makecell[l]{\textbf{MINT} \\ \textbf{(2026)}} & 
\textbf{[--] Low Risk:} Operates as a generative, preference-aligned fine-tuned LLM  that directly predicts diagnostic rankings without relying on case-based retrieval modules or external reference patient candidate banks, avoiding candidate-space leakage [p.~4; p.~15, \textit{Preference Optimization Frameworks}]. & 
\textbf{[+] High Risk:} Fine-tuned on GMDB ($n=6,522$), achieving 52.99\% Top-10 / 43.49\% Top-1 accuracy on internal test data ($n=386$) [p.~6; p.~32, \textit{Table~S1}]. When tested on the external Phenopacket disjoint subset ($n=4,342$; 396 unseen diseases), Top-10 and Top-1 accuracies plummeted to 10.48\% and 7.00\%—substantially underperforming even the base un-tuned LLM (20.00\% and 11.29\%) due to severe distribution overfitting [p.~8; p.~30, \textit{Table~2}; p.~38, \textit{Table~S5}]. & 
\textbf{[--] Low Risk:} Inputs are strictly constructed by concatenating standardized HPO terms with basic demographics (Age, Sex, Ethnicity), completely excluding retrospective clinical narratives, downstream therapeutic interventions, and discharge outcomes [p.~6; p.~14--15, \textit{SFT Data Construction}; p.~36--37, \textit{Table~S4}]. & 
\textbf{[+] High Risk:} Preference signals are generated using upstream GestaltMML (GMDB v1.0.9) and compared against specialized domain baselines (MedGemma-1.5) and RAG modules without clearly specifying the underlying KB versions (e.g., HPO and OMIM) or enforcing synchronized chronological KB snapshots across baselines [p.~6; p.~15, \textit{RAG}; p.~40, \textit{Table~S7}].\\

\makecell[l]{\textbf{RareSeek} \\ \textbf{(2025)}} & 
\textbf{[--] Low Risk:} Uses GraphRAG over a biomedical knowledge graph (ClinVar, HGMD, HPO, OMIM, Orphanet) via Cypher queries rather than case-based patient-similarity retrieval or reference patient candidate banks, avoiding patient-level candidate-space contamination [p.~5; p.~33--34, \textit{Knowledge Graph construction and RAG}]. & 
\textbf{[+] High Risk:} Fine-tuned on 48,852 EHRs from specific institutions (GWCMC and Tianjin platform), achieving 68.4\% Top-1 accuracy on internal test EHRs ($n=4,306$), but experienced a sharp drop to 39.2\% on the independent external RareBench dataset ($n=1,197$), and further fell to 19.2\% under pure phenotype inputs, demonstrating strong institutional overfitting [p.~6; p.~8--9, \textit{Fig.~2a--d}; p.~26; p.~30]. & 
\textbf{[+] High Risk:} Directly processes full retrospective EHR narratives including discharge summaries. Authors explicitly confirm that non-phenotypic evidence accounted for a median of 23.1\% of reasoning features, with 23.0\% stemming from post-hoc clinical interventions/procedures (e.g., Sildenafil therapy) and therapeutic responses, which are unavailable during prospective initial visits [p.~2; p.~9; p.~21, \textit{Fig.~5b}; p.~26, \textit{Methods}]. & 
\textbf{[+] High Risk:} GraphRAG incorporates contemporary databases (ClinVar 20250209, HGMD 2024.2) while benchmarking against traditional phenotype tools (Exomiser, Phen2Disease, PhenoDP, Phrank) operating on historically frozen ontology releases without chronologically synchronizing KB snapshots across baselines [p.~6; p.~33--35, \textit{Baselines for rare disease diagnosis}].\\

\makecell[l]{\textbf{Deep-DxSearch} \\ \textbf{(2025)}} & 
\textbf{[+] High Risk:} The patient retrieval database ($n=177,029$) was populated by randomly partitioning cases from the same evaluation sources (e.g., 419 RareBench cases and 46,037 RareArena/PMC-Patients cases were retained directly in the retrieval bank) without source-level isolation or PMID deduplication [p.~7; p.~28--29, \textit{Fig.~5}; p.~30, \textit{A.4.2}]. Ablation confirms that removing the patient database causes rare-disease Acc@1 to plummet from 55.20\% to 10.83\% [p.~35, \textit{Table~11}]. & 
\textbf{[+] High Risk:} The agentic policy model was trained via reinforcement learning (GRPO) predominantly on in-distribution (ID) intra-dataset splits across 6 centers, achieving high ID rare disease Acc@1 (70.48\% on RareBench) [p.~7; p.~8, \textit{Table~1}]. When tested on the independent out-of-distribution (OOD) cohort from Xinhua Hospital, diagnostic Acc@1 dropped sharply to 35.13\% (a $\sim$50\% drop), reflecting substantial distribution memorization [p.~9, \textit{Table~2}; p.~12, \textit{Table~3}]. & 
\textbf{[--] Low Risk:} Utilizes a multi-stage filtering and phenotype extraction pipeline to ensure case narratives reflect causal onset presentations rather than post-diagnostic sequelae, systematically stripping downstream pharmacological treatments, surgical interventions, and explicit diagnostic labels [p.~29--30, \textit{A.3 Diagnostic Data Processing}]. & 
\textbf{[+] High Risk:} Integrates up-to-date disease guidelines (16,371 conditions) and 23.9M PubMed articles while evaluating against diverse baselines (MedCPT 2023, CoD 2024, MedGemma 2025, Baichuan-M1 2025) without explicitly standardizing or synchronizing reference KB snapshot versions across all compared systems [p.~7; p.~10; p.~21, \textit{Baselines}].\\

\makecell[l]{\textbf{MEDDxAgent} \\ \textbf{(2025)}} & 
\textbf{[+] High Risk:} The \textit{Diagnosis Strategy Agent} employs dynamic few-shot retrieval (via BioClinicalBERT/BGE embeddings) to retrieve Top-$5$ similar reference cases directly from the evaluation datasets (DDxPlus, RareBench) without source-level isolation or PMID deduplication [p.~3; p.~6, \textit{Hyperparameters}]. Table~1 demonstrates that dynamic few-shot matching inflates GTPA@1 on RareBench from 0.46 (Zero-shot) to 0.82 (Few-shot CoT Dyn\_BAII) on GPT-4o, confirming severe performance inflation via candidate-space leakage [p.~7, \textit{Table~1}; p.~23, \textit{Table~6}]. & 
\textbf{[--] Low Risk:} Operates as a training-free agent framework orchestrating frozen foundation LLMs (GPT-4o, Llama-3.1-70B/8B) without supervised parameter fine-tuning on institutional clinical EHRs, avoiding parameter-level institutional distribution memorization [p.~2, \textit{MEDDxAgent Overview}; p.~5, \textit{Models and Tasks}]. & 
\textbf{[--] Low Risk:} For rare diseases (RareBench), diagnostic profiles are strictly structured into standardized symptom/phenotype lists, completely excluding retrospective clinical narratives, downstream therapeutic drug interventions, and discharge outcomes [p.~5, \textit{Benchmark Compilation}; p.~16, \textit{RareBench Profile}]. & 
\textbf{[+] High Risk:} The \textit{Knowledge Retrieval Agent} dynamically retrieves live online PubMed and Wikipedia articles at runtime via APIs without establishing publication date cutoff controls or chronologically synchronizing reference KB snapshot versions across compared foundation models [p.~3, \textit{Knowledge Retrieval Agent}; p.~6; p.~22, \textit{Table~5}].\\

\makecell[l]{\textbf{RareScale} \\ \textbf{(2025)}} & 
\textbf{[+] High Risk:} The candidate generation model (Llama-3.1-7B) is fine-tuned on synthetic dialogues derived from the expert system, and its predicted Top-$5$ candidates are explicitly injected into the prompt of the downstream black-box LLMs [p.~5, \textit{Diagnosis Generation}; p.~20, \textit{Prompt~3}]. Both candidate generation and final evaluation share the exact same closed-world synthetic rules (575 diseases), creating an artificial candidate-space shortcut [p.~3, \textit{Corpus Simulation}; p.~5, \textit{Table~1}]. & 
\textbf{[+] High Risk:} Trained and evaluated exclusively on synthetic patient chats simulated by specific LLMs (GPT-4o and Claude-3.5-Sonnet) using an internal QMR expert system [p.~3--4; p.~12, \textit{A.1, A.2}]. Table~3 shows severe distribution shift: a candidate model trained solely on GPT-4o chats achieves 88.04\% Top-5 accuracy on GPT-4o test chats, but collapses to 44.18\% on Claude test chats (a $\sim$50\% drop), and the study completely lacks evaluation on independent real-world clinical EHRs [p.~7, \textit{Table~3}; p.~8, \textit{Limitations}]. & 
\textbf{[--] Low Risk:} The chat simulation prompts explicitly instruct the model to simulate only initial history-taking encounters focusing on patient-reported symptoms, strictly excluding downstream laboratory results, imaging, confirmatory procedures, and therapeutic drug treatments [p.~3, \textit{Generation of structured cases}; p.~18--19, \textit{Prompt~1 \& Prompt~2}]. & 
\textbf{[?] Unclear:} Diagnostic rules are derived from legacy Internist-1/QMR expert systems, whereas candidate ranking and final DDx rely on modern foundation LLMs (GPT-4o, Claude-3.5-Sonnet, Llama-3.3-70B) without reporting temporal version synchronization across underlying expert rules and LLM pre-training cutoffs [p.~3; p.~5, \textit{Diagnosis Generation}]. \\

\makecell[l]{\textbf{MAC} \\ \textbf{(2025)}} & 
\textbf{[--] Low Risk:} Operates strictly as a multi-agent conversational debate framework (Admin, Doctor, and Supervisor agents via AutoGen) without incorporating a reference-patient retrieval database or dynamic few-shot case retrieval, eliminating candidate-space retrieval contamination [p.~3, \textit{Fig.~2}; p.~10, \textit{Multi-agent conversation framework}]. & 
\textbf{[--] Low Risk:} Built entirely as a training-free prompt-orchestration system over frozen foundation LLMs (GPT-3.5, GPT-4, GPT-4o-mini) without supervised parameter fine-tuning on institutional clinical EHRs, avoiding parameter-level institutional distribution memorization [p.~1, \textit{Abstract}; p.~10, \textit{Study design}]. & 
\textbf{[+] High Risk:} Diagnostic inputs directly utilize retrospective published case reports ($n=302$). As shown in Fig.~3, the text incorporates detailed pre-visit therapeutic interventions (e.g., \textit{``He was treated with anti-infective treatment... received cefotaxime before the visit''}) and pathognomonic diagnostic imaging features (e.g., \textit{``Both CT and MRCP showed obvious central spot sign''}), embedding rich retrospective diagnostic clues that are unavailable during prospective encounters [p.~4, \textit{Fig.~3a, b}; p.~10, \textit{Data extraction}]. & 
\textbf{[--] Low Risk:} The framework relies entirely on the internal parameterized reasoning of foundation LLMs without coupling to external evolving KBs; all compared baselines (GPT-3.5/GPT-4, CoT, Self-Refine, Self-Consistency) are evaluated under identical LLM versions without ontology version disparities [p.~3; p.~10, \textit{Methods}; p.~12, \textit{Comparison with other methods}]. \\

\makecell[l]{\textbf{RDguru} \\ \textbf{(2025)}} & 
\textbf{[--] Low Risk:} The multi-source fusion model (MixDiagDQN) was trained exclusively on 10,000 simulated cases synthesized from Orphadata and tested on 238 published cases, without utilizing a reference-patient retrieval database or patient-similarity matching, avoiding candidate-space patient leakage [p.~4--5, \textit{Training and Testing Data}; p.~11, \textit{MixDiagDQN Training}]. & 
\textbf{[--] Low Risk:} Uses frozen foundation LLMs (GPT-3.5-turbo/GPT-4) without supervised parameter fine-tuning on institutional EHRs. The DQN agent coordinates rank fusion based on synthetic case simulations (covering 936 RDs) rather than memorizing single-center clinical distributions [p.~5, \textit{Algorithm~1}; p.~11, \textit{Discussion}]. & 
\textbf{[--] Low Risk:} Clinical narratives are converted into standardized HPO terms via NCBO Annotator and FastContext. Diagnostic reasoning (PheLR, phenotype matching, RDmaster Q\&A) operates strictly over structured HPOs, structurally filtering out retrospective treatment courses, invasive procedures, and discharge summaries [p.~3--4, \textit{Phenotype Annotation}; p.~8, \textit{Fig.~3b}]. & 
\textbf{[?] Unclear:} Diagnostic fusion integrates static Orphadata-based computational models (PheLR) with GPT-4 without explicitly reporting chronological synchronization between KB snapshots and the publication dates of evaluated historical cases [p.~3, \textit{RD Knowledge Q\&A}; p.~5]. \\

\makecell[l]{\textbf{RareBench} \\ \textbf{(2024)}} & 
\textbf{[+] High Risk:} Dynamic few-shot prompting generates patient embeddings via IC-based random walks and retrieves Top-$m$ similar reference cases directly from the evaluation benchmark ($n=2,185$, combining public and PUMCH cohorts) without dataset-level isolation or PMID deduplication [p.~4--6, \textit{Sec.~3.4.3}; p.~12, \textit{A.3.4}]. Authors explicitly noted concerns that similar case injection could bias diagnostic results, leading them to avoid few-shot setups in certain tasks [p.~6, \textit{Sec.~4.2.2}; p.~7, \textit{Sec.~4.3.2}]. & 
\textbf{[--] Low Risk:} Primarily benchmarks in-context learning (zero-shot, CoT, few-shot) of 11 existing foundation LLMs without conducting supervised parameter fine-tuning on institutional clinical EHRs, avoiding parameter-level institutional distribution memorization [p.~4, \textit{Table~2}; p.~6, \textit{Table~4}]. & 
\textbf{[+] High Risk:} In Task 1, 2, 3, and the human--LLM comparison in Task 4, direct unstructured EHR narratives are supplied. As shown in the prompt example, the raw text incorporates detailed objective findings from prior hospitalization (color echocardiography) and explicit outpatient diagnostic impressions (e.g., \textit{``color echocardiography: tricuspid valve disease... admitted to our hospital for outpatient consideration of tricuspid valve disease''}), embedding decisive retrospective diagnostic clues [p.~8, \textit{Sec.~5.2.2, Table~6}; p.~11, \textit{A.3.1}]. & 
\textbf{[?] Unclear:} Constructed knowledge graphs fix HPO to release 2023-06-06 and CCRD to 2019-11 [p.~5, \textit{Table~3}; p.~12, \textit{A.4.1}]. Historical cohorts (RAMEDIS 2006, MME 2015, HMS 2019, LIRICAL 2020) are evaluated against modern 2023 knowledge graphs (introducing temporal backfeeding), and 11 LLMs with disparate pre-training knowledge cutoff dates are evaluated without chronological synchronization [p.~4, \textit{Table~2}; p.~6, \textit{Table~4}].\\

\makecell[l]{\textbf{PhenoBrain} \\ \textbf{(2025)}} & 
\textbf{[--] Low Risk:} Operates as an ensemble of ontology-driven Bayesian models (ICTO, PPO) and few-shot classifiers (CNB, MLP) without deploying a reference-patient retrieval database or patient-similarity matching, eliminating candidate-space retrieval contamination [p.~2--3, \textit{Overview}; p.~10--12, \textit{Diagnostic methods}]. & 
\textbf{[--] Low Risk:} Classifiers are trained solely on synthetic representations augmented from KB disease annotations (OMIM, Orphanet, CCRD via Mixup/perturbation) rather than fitting single-center EHR weights [p.~11--12]. Evaluated across diverse multi-country public datasets ($n=873$) and real-world clinical EHRs ($n=1,063$), maintaining consistent Top-10 recall (0.640 on public vs. 0.630 on PUMCH-L) without OOD collapse [p.~3--5, \textit{Table~1, Table~2}]. & 
\textbf{[--] Low Risk:} Clinical narratives are automatically parsed into standardized HPO terms via PBTagger; diagnostic models reason strictly over structured HPO vectors [p.~2, \textit{Fig.~1}; p.~12]. Clinical texts strictly redacted disease names/abbreviations, disease-specific diagnostic test descriptions, and genetic results, preventing post-hoc therapeutic or administrative clue leakage [p.~10, \textit{Data description}]. & 
\textbf{[--] Low Risk:} Integrates an explicitly version-controlled knowledge base (HPO as of 2023-04-05, CHPO 2022-10-05, OMIM, Orphanet, CCRD; 9,260 diseases) [p.~9, \textit{Methods}]. All 12 reproduced benchmark methods (BOQA, MinIC, Res, Lin, etc.) were evaluated on the exact same synchronized KB and datasets under identical evaluation criteria [p.~4, \textit{Performance evaluation}; p.~12]. \\

\makecell[l]{\textbf{PhenoDP} \\ \textbf{(2025)}} & 
\textbf{[--] Low Risk:} The diagnostic \textit{Ranker} combines IC-based, phi-based, and GCN semantic similarity measures directly against KB disease profiles (4,731 OMIM and 3,654 Orphanet entries) without deploying a reference-patient retrieval database or patient-similarity matching, eliminating candidate-space retrieval contamination [p.~3--4, \textit{The Ranker module}; p.~8, \textit{Fig.~1c}]. & 
\textbf{[--] Low Risk:} The GCN encoder is pre-trained via unsupervised DAG reconstruction and contrastive pairs are synthesized directly from public KBs rather than fitting single-center EHR weights [p.~4, \textit{Semantic-based similarity}]. Evaluated across simulated datasets ($n=3,000$) and three multi-source real-world cohorts (LIRICAL $n=381$, curated literature $n=130$, updated LIRICAL $n=5,485$), maintaining consistent superior ranking performance without OOD collapse [p.~5; p.~10--13, \textit{Fig.~4, Fig.~5}]. & 
\textbf{[--] Low Risk:} Diagnostic reasoning in the \textit{Ranker} and symptom recommendation in the \textit{Recommender} operate strictly on sets of standardized HPO terms ($Q = \{h_1, \dots, h_n\}$), structurally excluding retrospective unstructured text, therapeutic treatments, and discharge outcomes from diagnostic inputs [p.~3, \textit{The Ranker module}; p.~5, \textit{Real-world patient datasets}]. & 
\textbf{[--] Low Risk:} Explicitly documents and fixes reference ontology versions (OMIM April 17, 2024; Orphanet April 18, 2024; HPO April 26, 2024) [p.~3; p.~18, \textit{Data availability}]. All seven competing baseline methods (PhenoPro, GDDP, Phen2Disease, Phrank, LIRICAL, Phenomizer, Exomiser) were benchmarked locally under identical input phenotype sets and standardized metrics [p.~6--7, \textit{Competing methods}; p.~10]. \\

\makecell[l]{\textbf{PheLR} \\ \textbf{(2023)}} & 
\textbf{[--] Low Risk:} Implements a statistical likelihood ratio model derived from HPO DAG annotation propagation rules and Orphanet epidemiology priors, calculating analytical disease posterior probabilities directly against KB profiles (4,257 ORPHA diseases) without deploying a reference-patient retrieval database or patient-similarity matching, eliminating candidate-space contamination [p.~2--4, \textit{Estimating likelihood ratio}; p.~4, \textit{Evaluation}]. & 
\textbf{[--] Low Risk:} Operates as a purely mathematical, training-free probabilistic reasoning algorithm without supervised parameter updates or fine-tuning on institutional clinical EHRs [p.~3--4, \textit{Material and methods}]. Evaluated across extensive controlled simulations ($n=120,000$) and independent real-world published cases ($n=180$), exhibiting robust generalization against noise without institutional distribution collapse [p.~5--7, \textit{Results}; p.~6, \textit{Fig.~3}; p.~7, \textit{Fig.~5}]. & 
\textbf{[--] Low Risk:} Real-world case reports were parsed via HpoTextMining and manually proofread into standard Phenopackets containing only verified present and absent HPO terms; diagnostic reasoning is executed strictly over structured HPO inputs, structurally excluding retrospective clinical narratives, downstream treatments, and discharge outcomes [p.~4, \textit{Evaluation}; p.~7, \textit{Benchmarking performance}]. & 
\textbf{[--] Low Risk:} Explicitly documents and fixes reference data versions: HPO ontology (\texttt{hp.obo}, 2022-02-14), HPO annotations (\texttt{phenotype.hpoa}, 2022-02-14), and Orphanet epidemiology data (2022-01-26) [p.~2, \textit{Data sources}]. All competing tools (LIRICAL, PhenoPro, BOQA, Phenomizer) were benchmarked locally under the exact same synchronized KB snapshots [p.~4, \textit{Evaluation}]. \\

\makecell[l]{\textbf{LIRICAL} \\ \textbf{(2020)}} & 
\textbf{[--] Low Risk:} Implements an analytical Bayesian likelihood ratio (LR) framework directly against curated computational disease models in HPO/OMIM/Orphanet (7,623 diseases) without deploying a reference-patient retrieval database or patient-similarity matching, eliminating candidate-space retrieval contamination [p.~2, \textit{Material and Methods}; p.~6, \textit{Ranking Candidates}]. & 
\textbf{[--] Low Risk:} Operates as a purely mathematical, training-free probabilistic reasoning algorithm without supervised parameter updates or fine-tuning on institutional clinical EHRs [p.~2; p.~4--5, \textit{Genotype LR}]. Evaluated across 384 curated literature cases (262 Mendelian diseases) and 116 independent solved singleton cases from the 100,000 Genomes Project across 89 genes without institutional distribution collapse [p.~6, \textit{Evaluation}; p.~10, \textit{Fig.~4}]. & 
\textbf{[--] Low Risk:} Diagnostic inputs strictly consist of standardized HPO terms (present and explicitly excluded phenotypes) and VCF variant calls formatted in GA4GH Phenopackets, structurally excluding retrospective unstructured narratives, clinical treatments, and discharge summaries from the reasoning engine [p.~2, \textit{LR for Phenotypes}; p.~6, \textit{Evaluation}; p.~7, \textit{Table~1}]. & 
\textbf{[--] Low Risk:} Explicitly specifies and fixes reference database and ontology versions: HPO ontology (\texttt{hp.obo}, 2019-09-06), HPO annotations (\texttt{phenotype.hpoa}, 2019-10-16), Exomiser database (v12.1.0), and gnomAD population frequencies [p.~2, \textit{Data Sources}; p.~4, \textit{Exomiser pathogenicity score}]. All comparative evaluations against Phenomizer and Exomiser were benchmarked under identical synchronized configurations [p.~6; p.~9--10, \textit{Fig.~2}]. \\

\makecell[l]{\textbf{BOQA} \\ \textbf{(2012)}} & 
\textbf{[--] Low Risk:} Formulates disease prioritization via a three-layered Bayesian network to analytically infer marginal posterior probabilities $P(I_j \mid Q)$ across 2,368 Mendelian diseases directly against HPO annotations, without deploying a reference-patient retrieval database or patient-similarity matching, eliminating candidate-space retrieval contamination [p.~2--4, \textit{Sec.~2.1--2.7, Algorithm~1}]. & 
\textbf{[--] Low Risk:} Operates as a purely mathematical, training-free graphical model parameterizing false-positive ($\alpha$) and false-negative ($\beta$) query errors without supervised parameter fine-tuning on institutional clinical EHRs [p.~3--4, \textit{Sec.~2.5, 2.8}]. Benchmarked across 11,840 simulated patient profiles over 2,368 diseases under varied noise levels, demonstrating robust statistical generalizability without institutional distribution collapse [p.~5--6, \textit{Sec.~2.10, 3, Table~1, Fig.~4}]. & 
\textbf{[--] Low Risk:} Diagnostic queries are strictly composed of structured HPO term sets representing observed clinical manifestations ($Q = \{Q_1, \dots, Q_m\}$), structurally excluding retrospective unstructured text, therapeutic drug interventions, and discharge outcomes from diagnostic inputs [p.~3, \textit{Sec.~2.2, 2.3}; p.~5, \textit{Sec.~2.10}]. & 
\textbf{[--] Low Risk:} Explicitly documents and fixes the reference ontology and annotation snapshot downloaded on June 1, 2011 (2,368 diseases, 6,584 HPO terms) [p.~5, \textit{Sec.~2.10}]. All four competing baseline methods (Res, Lin, JC, and ResP/Phenomizer) were evaluated on the exact same synchronized HPO snapshot and test queries under identical conditions [p.~5--6, \textit{Sec.~2.10, Fig.~4}]. \\

\makecell[l]{\textbf{Phenomizer} \\ \textbf{(2009)}} & 
\textbf{[--] Low Risk:} Implements an ontological semantic similarity search algorithm (using information content of the most informative common ancestor, MICA) to compute similarity scores and statistical $p$-values directly against OMIM disease annotations (4,813 diseases) without utilizing a reference-patient retrieval database or patient-similarity matching, eliminating candidate-space retrieval contamination [p.~2--3, \textit{Material and Methods}; \textit{Eqs.~1--2}]. & 
\textbf{[--] Low Risk:} Operates as a purely mathematical, training-free semantic search and statistical ranking algorithm without supervised parameter updates or fine-tuning on institutional clinical EHRs [p.~2--4]. Validated using 4,400 simulated patient profiles across 44 complex dysmorphology syndromes under systematic permutations of clinical noise (randomly added terms) and imprecision (ancestor term replacement), avoiding institutional distribution memorization [p.~4--5, \textit{Performance Evaluation}; p.~6, \textit{Fig.~3}]. & 
\textbf{[--] Low Risk:} Diagnostic queries strictly consist of sets of standardized HPO terms representing observed phenotypic abnormalities, structurally excluding unstructured narrative clinical text, retrospective therapeutic treatments, invasive procedures, and discharge summaries from the search inputs [p.~2, \textit{Material and Methods}; p.~4, \textit{Results}]. & 
\textbf{[--] Low Risk:} Explicitly documents and fixes the reference ontology version (HPO release May 6, 2009, version 1.58 from NCBO Bioportal, ID 1125, containing $\sim$9,000 terms and $\sim$50,000 annotations across 4,813 OMIM diseases) [p.~2, \textit{Material and Methods}]. All compared baselines (Feature Vector method, one-sided OSS, symmetric OSS, OSS-PV) were benchmarked locally under the exact same synchronized ontology snapshot [p.~2--5]. \\

\end{xltabular}
\endgroup

\clearpage
\begin{table}[h]
\caption{\textbf{Case sources and accessibility.}}
\label{tab:dataset_url}
\renewcommand{\arraystretch}{1.2}
\begin{tabularx}{\textwidth}{l >{\centering\arraybackslash}X}
\toprule
\textbf{Dataset} & \textbf{Access URL} \\
\midrule
RareArena           & https://github.com/chen/rarearena \\
Phenopacket           & https://github.com/monarch-initiative/phenopacket-store \\
LIRICAL    & https://zenodo.org/record/3905420 \\
RDmaster            & https://github.com/1678492347/rdmaster-test \\
PhenoDP                & https://link.springer.com/article/10.1186/s13073-025-01496-8\#Sec37 \\
MME         & https://github.com/ga4gh/mme-apis/tree/master/testing \\
VPI               & https://www.nature.com/articles/ejhg2017123\#Sec9 \\
HMS                    & https://link.springer.com/article/10.1186/s13023-019-1040-6\#Sec37 \\
RAMEDIS              & http://www.ramedis.de \\
PUMCH          & https://zenodo.org/records/10774650 \\
ZJUMCH (in-house)                               & -- \\
\bottomrule
\end{tabularx}
\end{table}